# Parameterized Complexity Results for
# Exact Bayesian Network Structure Learning

**Sebastian Ordyniak**                                    ORDYNIAK@FI.MUNI.CZ
*Masaryk University Brno, Czech Republic*

**Stefan Szeider**                                        STEFAN@SZEIDER.NET
*Vienna University of Technology, Austria*

## Abstract

Bayesian network structure learning is the notoriously difficult problem of discovering a Bayesian network that optimally represents a given set of training data. In this paper we study the computational worst-case complexity of exact Bayesian network structure learning under graph theoretic restrictions on the (directed) super-structure. The super-structure is an undirected graph that contains as subgraphs the skeletons of solution networks. We introduce the directed super-structure as a natural generalization of its undirected counterpart. Our results apply to several variants of score-based Bayesian network structure learning where the score of a network decomposes into local scores of its nodes.

*Results:* We show that exact Bayesian network structure learning can be carried out in non-uniform polynomial time if the super-structure has bounded treewidth, and in linear time if in addition the super-structure has bounded maximum degree. Furthermore, we show that if the directed super-structure is acyclic, then exact Bayesian network structure learning can be carried out in quadratic time. We complement these positive results with a number of hardness results. We show that both restrictions (treewidth and degree) are essential and cannot be dropped without loosing uniform polynomial time tractability (subject to a complexity-theoretic assumption). Similarly, exact Bayesian network structure learning remains NP-hard for "almost acyclic" directed super-structures. Furthermore, we show that the restrictions remain essential if we do not search for a globally optimal network but aim to improve a given network by means of at most $k$ arc additions, arc deletions, or arc reversals ($k$-neighborhood local search).

## 1. Introduction

Bayesian Network Structure Learning (BNSL) is the important task of discovering a Bayesian network that represents a given set of training data. Unfortunately, solving the problem optimally (Exact BNSL) is NP-complete (Chickering, 1996). A common and widely used approach to overcome this complexity barrier is to exploit the structure of the problem. This has also been a popular direction for BNSL and two main kinds of structural restrictions have been studied so far, i.e., (1) restrictions on the probability distribution generating the input and (2) restrictions on the resulting Bayesian network. With the help of these restrictions several tractable classes of BNSL have been identified. BNSL is solvable in non-uniform polynomial time if the distribution generating the input has bounded treewidth (Narasimhan & Bilmes, 2004) or bounded degree (Pieter, Daphne, & Andrew, 2006) and it is solvable in (non-uniform) polynomial time if the resulting BN is a branching (Chow & Liu, 1968) or close to being a branching (Gaspers, Koivisto, Liedloff, Ordyniak,

& Szeider, 2012). These positive results are contrasted by a series of negative results for the above mentioned restrictions, e.g., BNSL is known to be NP-hard if the resulting BN is a polytree (Dasgupta, 1999) or a directed path (Meek, 2001). Recently, a novel approach to restrict the structure of BNSL has been introduced (Tsamardinos, Brown, & Aliferis, 2006; Perrier, Imoto, & Miyano, 2008). Here a so-called super-structure, an undirected graph on the same nodes as the resulting BN, is used to restrict the search space of BNSL in advance. After the super-structure has been obtained, usually using an IT-based approach (Tsamardinos et al., 2006), one looks for solution networks whose skeletons are contained in the super-structure. It hence becomes important that the super-structure is sound, i.e., contains at least one optimal solution.

There are two *main questions* concerning the super-structure: First, how can a suitable and sound super-structure be obtained efficiently? And secondly, once such a super-structure is obtained, how can it be used to guide the search for an optimal solution? The goal of this paper is to provide a theoretical analysis of the latter question, considering super-structures that arise from a given local score function adapting the model of Parviainen and Koivisto (2010). We consider various combinations of restrictions in a systematic way that allows us draw a broader picture of the complexity landscape of BNSL. We hope that our analysis can help to understand the boundaries between tractable and intractable cases of this important problem. Furthermore, we think that our results provide new insights that can help the search for more efficient and accurate heuristics. For our analysis, we use the theoretical framework of *parameterized complexity* (Downey & Fellows, 1999) which seems to be well suited for investigating the complexity of BNSL as it allows to take structural properties (in terms of parameters) into account. To the best of our knowledge parameterized complexity theory has not been employed in this context before.

## 1.1 Results

In this section we give a brief overview of our results.

### 1.1.1 EXACT BNSL USING THE SUPER-STRUCTURE

In the first part of our paper we study the worst-case complexity of Exact BNSL under graph-theoretic restrictions on the super-structure. One of the most prominent restrictions on the super-structure that we consider is treewidth. Treewidth is an important and widely used parameter that measures the similarity of a graph to a tree (Bodlaender, 1993, 1997, 2005; Greco & Scarcello, 2010). Similarly as for trees, many otherwise intractable problems become tractable on graphs of bounded treewidth. More importantly, treewidth has already been successfully applied in the context of Bayesian Reasoning (Darwiche, 2001; Dechter, 1999; Kwisthout, Bodlaender, & van der Gaag, 2010). It hence seems only natural to apply treewidth to (Exact) BNSL.
Our results are as follows:

(1) Exact BNSL is solvable in *non-uniform polynomial* time if the treewidth of the super-structure is bounded by an arbitrary constant.

(2) Exact BNSL is solvable in *linear time* if both treewidth and maximum degree of the super-structure are bounded by arbitrary constants.

By "non-uniform" we mean that the order of the polynomial depends on the treewidth. We obtain results (1) and (2) by means of a dynamic programming algorithm along a decomposition tree of the super-structure.

We show that—in a certain sense—both results are optimal:

(3) Exact BNSL for instances with super-structures of maximum degree 4 (but unbounded treewidth) is not solvable in polynomial time unless P = NP. Thus, in (1) and (2) we cannot drop the bound on the treewidth.

(4) Exact BNSL for instances with super-structures of bounded treewidth (but unbounded maximum degree) is not solvable in uniform polynomial time unless FPT = W[1]. Thus, in (2) we cannot drop the bound on the degree.

FPT ≠ W[1] is a widely accepted complexity theoretic assumption (Downey & Fellows, 1999) that is often considered as the parameterized analog to P ≠ NP. We will provide necessary background on parameterized complexity and fpt-reductions in Section 2.2.

### 1.1.2 Local Search BNSL Using the Super-structure

Since learning an optimal Bayesian network is computationally hard heuristic methods are used in practice. A popular heuristic for BNSL is the so-called hill climbing procedure, or local search. In particular, a highly competitive algorithm for learning large Bayesian networks (MMHC) uses local search to find an optimal solution inside a previously constructed super-structure (Tsamardinos et al., 2006). We study the worst case complexity of a well-known generalization of local search, the *k-Neighborhood Local Search* (or *k-Local Search* for short). In this variant of local search one is allowed to modify not one but up to $k$ arcs in every step of the search. Hence by adjusting $k$, one is able to balance speed and accuracy. However, because the $k$-local search space is of order $n^{O(k)}$, applying $k$-local search is especially desirable for problems where the running time does only increase modestly with respect to $k$. Similarly to result (2) we are able to show the following:

(5) $k$-Local Search BNSL is solvable in *linear time* if both the treewidth and the maximum degree of the super-structure are bounded by arbitrary constants.

Clearly, this result is only of minor interest because we can already solve the Exact BNSL problem under the same restrictions on the super-structure (see result (2)). However, in contrast to the Exact BNSL problem one might be able to drop one of these restrictions without losing uniform polynomial-time tractability. We show that this is again unlikely.

(6) $k$-Local Search BNSL for instances with super-structures that have either bounded treewidth or bounded maximum degree is not solvable in uniform polynomial time unless FPT = W[1].

### 1.1.3 Exact BNSL Using the Directed Super-structure

So far, one has only considered the super-structure as an undirected graph. We introduce the directed super-structure as a more expressive way to restrict the search space of solutions, i.e., once the directed super-structure has been fixed we restrict our search to solutions whose networks are contained in the directed super-structure. Again, we study the complexity of Exact BNSL with respect to certain restrictions on the directed super-structure and obtain the following result.

| Problem: Exact BNSL | | Complexity | Result |
|---|---|---|---|
| *Restrictions on Super-Structure* | | | |
| bounded treewidth | bounded max degree | linear time | Corollary 1 |
| bounded treewidth | – | in XP, W[1]-hard | Corollary 1, Theorem 3 |
| – | bounded max degree | NP-hard | Theorem 2 |
| *Restrictions on Directed Super-Structure* | | | |
| acyclic | – | poly-time | Corollary 2 |
| almost acyclic | bounded max degree | NP-hard | Theorem 7 |

| Problem: k-Local Search BNSL | | Complexity | Result |
|---|---|---|---|
| *Restrictions on Super-Structure* | | | |
| bounded treewidth | bounded max degree | linear time | Proposition 6 |
| bounded treewidth | – | in XP, W[1]-hard | Theorem 5 |
| – | bounded max degree | in XP, W[1]-hard | Theorem 6 |
| *Restrictions on Operations* | | | |
| only arc deletion | | in FPT | Theorem 4 |
| only arc addition | | in FPT | Theorem 4 |

Figure 1: Overview of complexity results. The complexity classes FPT and XP contain all problems solvable in uniform polynomial-time and non-uniform polynomial-time, respectively.

(7) Exact BNSL is solvable in *quadratic time* if the directed super-structure is acyclic.

The question arises whether we can extend this result by using a more general form of acyclicity. We show, however, that this is not the case.

(8) Exact BNSL is NP-hard for directed super-structures that can be made acyclic by deleting one node. Hardness even holds if the maximum in-degree and the maximum out-degree of the directed super-structure are both bounded by 3.

A systematic overview of our results (1)–(8) is given in Figure 1.

## 1.2 Related Work

In this section we present relevant related work on BNSL. Further related work can be found in the respective sections, i.e., we present related work on parameterized complexity in Section 2.2, related work on treewidth and tree decompositions in Sections 2.3 and 3.1, and in Section 7 we present related work on $k$-neighborhood local search.

### 1.2.1 Algorithms for Exact BNSL

To this date only a handful of exact algorithms for BNSL have been proposed. These can be split into three groups: (A) exact algorithms that do not employ any restrictions (Parviainen & Koivisto, 2010; Koivisto, 2006; Yuan, Malone, & Wu, 2011; Ott, Imoto, & Miyano, 2004; Silander & Myllymäki, 2006; Yuan et al., 2011), (B) exact algorithms using restrictions on the generating or target distribution (Pieter et al., 2006; Chechetka & Guestrin, 2007; Friedman, Nachman, & Pe'er, 1999; Chow & Liu, 1968; Gaspers, Koivisto et al., 2012), and (C) exact algorithms that use restrictions on the undirected super-structure (Friedman et al., 1999; Kojima, Perrier, Imoto, & Miyano, 2010). Algorithms falling into group (A) are only suited for small to medium sized Bayesian networks because they do not restrict the search space in any way. On the other hand, the restrictions used by algorithms in group (B) are more general than restrictions coming from the undirected super-structure and up to now only non-uniform polynomial-time algorithms could be obtained using these restrictions. To the best of our knowledge this paper is the first to employ an in-depth theoretical analysis of the parameterized complexity of BNSL using restrictions on the undirected super-structure. We obtain the first algorithm for exact BNSL with a uniform polynomial running-time with respect to structural restriction on the undirected super-structure. A similar approach has been taken by Kojima et al. (2010) where the authors propose an algorithm for exact BNSL that uses a "cluster-decomposition" of the undirected super-structure. Even though their practical results are quite promising the authors provide no theoretical analysis of the worst-case complexity of their algorithm beyond the trivial bound that also applies to exact algorithms in group (A). Apart from exact algorithms there also exists a variety of approximation algorithms using tree decomposition or degree-based techniques (Pieter et al., 2006; Elidan & Gould, 2008; Karger & Srebro, 2001).

### 1.2.2 Hardness Results for Exact BNSL

There are also a number of hardness results for BNSL under restrictions of the resulting Bayesian network. In particular, BNSL remains NP-hard if the in-degree of the resulting Bayesian network is bounded by 2 (Chickering, 1996), and if the resulting Bayesian network is a poly tree (Dasgupta, 1999), or a directed path (Meek, 2001). However, to the best of our knowledge, no negative results for BNSL under restrictions of the (directed) super-structure have been obtained.

## 1.3 Organization and Prior Work

This paper is organized as follows: In Section 2 we introduce the basic concepts and notions that we use throughout the paper. We introduce the main object of our study (BNSL) in Section 3. Section 4 shows how to use a dynamic programming algorithm on a tree decomposition in order to show results (1) and (2). We provide a refined complexity analysis of our algorithm in Section 5. In Section 6 we show the complexity boundaries for Exact BNSL using a super-structure, i.e., we obtain results (3) and (4). We introduce $k$-Local Search BNSL in Section 7 where we establish results (5) and (6). We introduce the directed super-structure in Section 8 and show results (7) and (8). We conclude in Section 9. An appendix contains the proofs of some technical claims.

A preliminary and shortened version of this paper appeared in the proceedings of UAI 2010. Apart from providing a higher level of detail and readability by giving more examples and detailed proofs, the paper extends its previous version in four ways: by adding a section on related work (Section 1.2), by providing a refined complexity analysis of our main algorithmic result (Section 5), by providing a novel proof of Theorem 6 that allows us to decrease the upper bound on the maximum degree of the super-structure from 5 to 3, and by introducing the directed super-structure (Section 8).

## 2. Preliminaries

In this section we will introduce the basic concepts and notions that we will use throughout the paper.

## 2.1 Basic Graph Theory

We will assume that the reader is familiar with basic graph theory (see, e.g., Diestel, 2000; Bang-Jensen & Gutin, 2009). We consider undirected graphs and directed graphs (digraphs). A *dag* is a directed acyclic graph. We write $V(G) = V$ and $E(G) = E$ for the sets of nodes and edges of a (directed or undirected) graph $G = (V, E)$, respectively. We denote an undirected edge between nodes $u$ and $v$ as $\{u, v\}$ and a directed edge (or arc), directed from $u$ to $v$ as $(u, v)$. We write $N_G(v)$ for the set of neighbors of a node $v \in V$ in $G$, i.e., $N_G(v) = \{ u : (v, u) \in E \text{ or } (u, v) \in E \}$ if $G$ is directed and $N_G(v) = \{ u : \{u, v\} \in E \}$ if $G$ is undirected. For a subset $V' \subseteq V$ we write $G[V']$ to denote the *induced subgraph* $G' = (V', E')$ where $E' = \{ e \subseteq V' : e \in E \}$ if $G$ is undirected and $E' = \{ e \in V' \times V' : e \in E \}$ if $G$ is directed. If $G$ is a digraph we define $\mathrm{P}_G(v) = \{ u \in V(G) : (u, v) \in E(G) \}$ as the set of *parents* of $v$ in $G$. Furthermore, for two directed graphs $D_1$ and $D_2$ we define $D_1 \cup D_2$ as the union of $D_1$ and $D_2$, i.e., $V(D_1 \cup D_2) = V(D_1) \cup V(D_2)$ and $E(D_1 \cup D_2) = E(D_1) \cup E(D_2)$. Let $G$ be a (directed or undirected) graph and $e \in E(G)$ a directed or undirected edge in $G$. We denote by $G - e$ the (undirected or directed) graph, where $G - e = (V(G), E(G) \setminus \{e\})$. Furthermore, for a subset $X \subseteq V(G)$ we denote by $G - X$ the graph induced by the nodes in $V \setminus X$. If $X$ contains only one node $v$ we also write $G - v$ instead of $G - \{v\}$. We call an undirected graph $G' = (V', E')$ the *skeleton* of a directed graph $G$ if $V' = V(G)$ and $E' = \{ \{u, v\} : (u, v) \in E(G) \}$.

## 2.2 Parameterized Complexity

Parameterized complexity provides a theoretical framework to distinguish between uniform and non-uniform polynomial-time tractability with respect to a parameter. It has been introduced and pioneered by Downey and Fellows (1999) and is receiving growing interest as reflected by the recent publication of two further monographs (Flum & Grohe, 2006; Niedermeier, 2006) and hundreds of research papers (see references in the above mentioned monographs). In 2008 the *Computer Journal* has devoted two special issues on parameterized complexity in order to make the key methods and ideas known to a wide range of computer scientists (Downey, Fellows, & Langston, 2008).

An instance of a parameterized problem is a pair $(I, k)$ where $I$ is the *main part* and $k$ is the *parameter*; the latter is usually a non-negative integer. A parameterized problem is *fixed-parameter tractable* if there exist a computable function $f$ and a constant $c$ such that instances $(I, k)$ of size $n$ can be solved in time $O(f(k)n^c)$. FPT is the class of all fixed-parameter tractable decision problems. Fixed-parameter tractable problems are also called *uniform polynomial-time tractable* because if $k$ is considered constant, then instances with parameter $k$ can be solved in polynomial time where the order of the polynomial is independent of $k$ (in contrast to non-uniform polynomial-time running times such as $n^k$).

Parameterized complexity offers a completeness theory similar to the theory of NP-completeness. One uses *fpt-reductions* which are many-one reductions where the parameter for one problem maps into the parameter for the other. More specifically, problem $L$ reduces to problem $L'$ if there is a mapping $R$ from instances of $L$ to instances of $L'$ such that (i) $(I, k)$ is a yes-instance of $L$ if and only if $(I', k') = R(I, k)$ is a yes-instance of $L'$, (ii) $k' \le g(k)$ for a computable function $g$, and (iii) $R$ can be computed in time $O(f(k)n^c)$ where $f$ is a computable function, $c$ is a constant, and $n$ denotes the size of $(I, k)$. The parameterized complexity class W[1] is considered as the parameterized analog to NP. In particular, FPT = W[1] implies the (unlikely) existence of a $2^{o(n)}$ algorithm for $n$-variable 3SAT (Impagliazzo, Paturi, & Zane, 2001; Flum & Grohe, 2006). An example, for a parameterized problem that is W[1]-complete under fpt-reductions is the parameterized Maximum Clique problem (given a graph $G$ and a parameter $k \ge 0$, does $G$ contain a complete subgraph on $k$ nodes?). Note that there exists a trivial non-uniform polynomial-time $n^k$ algorithm for the Maximum Clique problems that checks all sets of $k$ nodes.

## 2.3 Tree Decompositions

Treewidth is an important graph parameter that indicates in a certain sense the "tree-likeness" of an undirected graph (Bodlaender, 1993, 1997, 2005). On graphs of treewidth bounded by a constant many otherwise intractable problems become tractable. Bucket

Elimination (Dechter, 1999) and Recursive Conditioning (Darwiche, 2001) are two important algorithmic concepts that apply to instances of bounded treewidth.

The treewidth of a graph $G = (V, E)$ is defined via the following notion of decomposition: a *tree decomposition* of $G$ is a pair $(T, \chi)$ where $T$ is a tree and $\chi$ is a labeling function with $\chi(t) \subseteq V$ for every tree node $t$ such that the following conditions hold:

1. Every node of $G$ occurs in $\chi(t)$ for some tree node $t$.

2. For every edge $\{u, v\}$ of $G$ there is a tree node $t$ such that $u, v \in \chi(t)$.

3. For every node $v$ of $G$, let $T_v$ be the subgraph of $T$ induced by all nodes $t$ such that $v \in \chi(t)$. Then $T_v$ is a (connected) subtree of $T$ ("Connectedness Condition").

The *width* of a tree decomposition $(T, \chi)$ is the size of a largest set $\chi(t)$ minus 1 among all nodes $t$ of $T$. A tree decomposition of smallest width is *optimal*. The *treewidth* of a graph $G$, denoted tw$(G)$, is the width of an optimal tree decomposition of $G$. It is known that if $(T, \chi)$ is a tree decomposition of a graph $G$, then every clique of $G$ is contained in $\chi(t)$ for some tree node $t \in V(T)$ (Kloks, 1994, Lemma 2.2.2).

The following proposition will be useful to retrieve an upper bound on the treewidth of a graph.

**Proposition 1.** *Let $G$ be an undirected graph and $X \subseteq V(G)$. If the graph $G - X$ contains no edge, i.e., all the nodes in $G - X$ are isolated, then* tw$(G) \leq |X|$.

*Proof.* Let $V(G) \setminus X$ contain the nodes $v_1, \ldots, v_n$ and let $T$ be the tree with node set $\{t, t_1, \ldots, t_n\}$ and edge set $\{\{t, t_i\} : 1 \leq i \leq n\}$. Then $T$ together with the function $\chi$ such that $\chi(t) = X$ and $\chi(t_i) = X \cup \{v_i\}$ for every $1 \leq i \leq n$ is a tree decomposition for $G$ of width $|X|$. □

The main property of tree decompositions that allows for efficient bottom-up dynamic programming algorithms for a wide spectrum of otherwise intractable problems is its well-known separation property which is made precise in the following proposition.

**Proposition 2.** *Let $G$ be a graph, $(T, \chi)$ a tree decomposition for $G$, $\{t', t''\}$ be an edge in $T$, and let $T'$ and $T''$ be the subtrees of $T$ obtained from $T$ after deleting the edge $\{t', t''\}$ such that $T'$ contains $t'$ and $T''$ contains $t''$. Furthermore, define $S = \chi(t') \cap \chi(t'')$, $A = \bigcup_{t \in V(T')} \chi(t)$ and $B = \bigcup_{t \in V(T'')} \chi(t)$. Then the following two statements hold:*

1. $A \cap B = S$;

2. *$S$ separates the nodes in $A$ from the nodes in $B$, i.e., there is no edge between a node in $A \setminus S$ and a node in $B \setminus S$.*

*Proof.* The first statement follows immediately from Property (3) of a tree decomposition, since the subtree containing a node $v \in (A \cap B)$ has to make use of the edge $\{t', t''\}$ in order to be connected and hence $v$ has to be contained in $S$.

To see the second statement, suppose for a contradiction that there is an edge between a node $a \in A \setminus S$ and a node $b \in B \setminus S$. It follows from the first statement of this proposition that $T_a$ and $T_b$ are disjoint, i.e., otherwise either $a \in S$ or $b \in S$. But this contradicts property (2) of a tree decomposition since $\{a, b\}$ is an edge of $G$. □

Given a graph $G$ with $n$ nodes and a constant $w$, it is possible to decide whether $G$ has treewidth at most $w$, and if so, to compute an optimal tree decomposition of $G$ in time $O(n)$ (Bodlaender, 1996). Furthermore there exist powerful heuristics to compute tree decompositions of small width in a practically feasible way (Koster, Bodlaender, & van Hoesel, 2001; Gogate & Dechter, 2004; Dow & Korf, 2007). Recently, new randomized heuristics have been studied in the context of Bayesian reasoning (Kask, Gelfand, Otten, & Dechter, 2011; Gelfand, Kask, & Dechter, 2011).

## 3. Bayesian Network Structure Learning

In this section we define the theoretical framework for BNSL that we shall use for our considerations. We closely follow the abstract framework used by Parviainen and Koivisto (2010) which encloses a wide range of score-based approaches to structure learning. We assume that the input data specifies a set $N$ of *nodes* (representing random variables) and a *local score function* $f$ that assigns to each $v \in N$ and each subset $P \subseteq N \setminus \{v\}$ a non-negative real number $f(v, P)$. Given the local score function $f$ and a set $N$ of nodes, the problem is to find a dag $D$ on $N$ such that the *score* of $D$ under $f$

$$f(D) := \sum_{v \in N} f(v, \mathrm{P}_D(v))$$

is as large as possible (the dag $D$ together with certain local probability distributions forms a Bayesian network). This setting accommodates several popular scores like BDe, BIC and AIC (Parviainen & Koivisto, 2010; Chickering, 1995).

We consider the following decision problem:

---

Exact Bayesian Network Structure Learning

Input:     A local score function $f$ defined on a set $N$ of nodes, a real number $s > 0$.

Question:  Is there a dag $D$ on $N$ such that $f(D) \geq s$?

---

For the following complexity results we need to fix how the score function is represented in the problem input. We cannot list the values $f(v, P)$ for all nodes $v \in N$ and all subsets $P \subseteq N \setminus \{v\}$, as this requires $\Omega(2^{|N|})$ space. Therefore, we assume that $f(v, P)$ is only given if it is different from 0; we call this the *non-zero* representation. This representation is also used in other fields, for instance in constraint satisfaction, where only allowed tuples of constraints are listed in the input (Tsang, 1993). An alternative representation of the score function would be to list in the input all values $f(v, P)$ where $|P| \leq c$ for some constant $c$. Let us call this the *arity-c representation*. This representation requires only polynomial space if every node has at most $c$ parents, for a small constant $c$, which is a reasonable assumption. Given an arity-$c$ representation of a score function, we can clearly compute in linear time the corresponding non-zero representation. Hence the non-zero representation is more general, and therefore we will base our complexity results on this encoding. All tractability results also carry over to the arity-$c$ representation.

The *size* of $f$ (given in the non-zero representation) is the number of bits needed to represent the tuples with $f(v, P) > 0$ in a reasonable data structure, e.g., as a list of these tuples where each tuple stores the node, the set of parents, and the value of the score function. Clearly, the size of $f$ exceeds the total number of all such tuples. We define $\mathcal{P}_f(v) := \{ P \subseteq N : f(v, P) > 0\} \cup \{\emptyset\}$ to be the set of all *potential parent sets* of $v$. We also define

$$\delta_f := \max_{v \in N} |\mathcal{P}_f(v)|;$$

which will be an important measurement for our worst-case analysis of running times. In particular, the above assumption on how $f$ is represented implies the following:

(*) Let $I = (N, f, s)$ be an instance of Exact Bayesian Network Structure Learning. Then $\delta_f$ is bounded by the size of $f$.

Let $f$ be a local score function defined on a set $N$ of nodes. The *directed super-structure* of $f$ is the directed graph $S_f^{\rightarrow} = (N, E_f)$ where $E_f$ contains an arc $(u, v)$ if and only if $u$ is a potential parent of $v$, i.e., if $u \in P$ for some $P \in \mathcal{P}_f(v)$. The (undirected) *super-structure* of $f$, denoted $S_f = (N, E_f)$, is the skeleton of the directed super-structure.

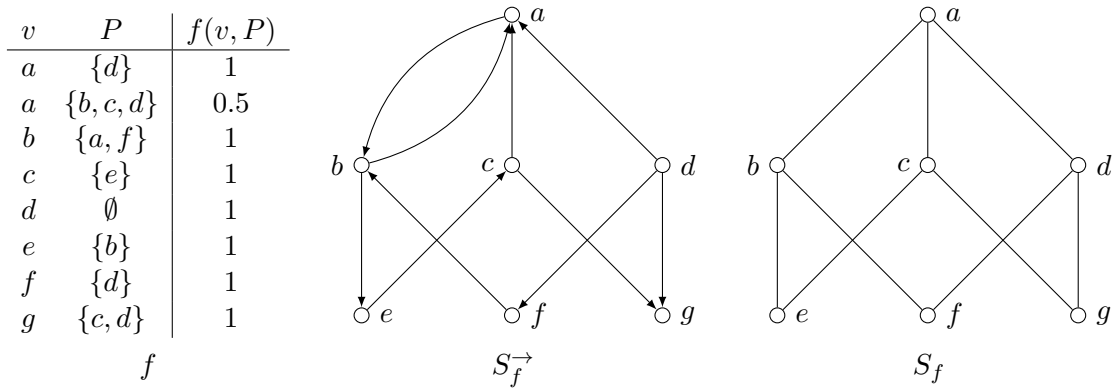

| $v$ | $P$ | $f(v, P)$ |
|-----|-----|-----------|
| $a$ | $\{d\}$ | 1 |
| $a$ | $\{b, c, d\}$ | 0.5 |
| $b$ | $\{a, f\}$ | 1 |
| $c$ | $\{e\}$ | 1 |
| $d$ | $\emptyset$ | 1 |
| $e$ | $\{b\}$ | 1 |
| $f$ | $\{d\}$ | 1 |
| $g$ | $\{c, d\}$ | 1 |

$f$ $\qquad\qquad S_f^{\rightarrow} \qquad\qquad S_f$

Figure 2: A local score function $f$ together with the directed super-structure $S_f^{\rightarrow}$ and the undirected super-structure $S_f$.

**Example 1.** *Figure 2 shows an example for a local-score function $f$ defined on the set $N = \{a, b, c, d, e, f, g\}$ of nodes. The function $f$ is given as a table containing all tuples $(v, P, f(v, P))$ with $v \in N$ and $P \subseteq N \setminus \{v\}$ such that $f(v, P) > 0$. Note that since $f$ is non-negative it holds that $f(v, P) = 0$ for all the remaining pairs $(v, P)$, i.e., the pairs that are not contained in the table. The figure also shows the directed super-structure $S_f^{\rightarrow}$ and the unique super-structure $S_f$ of $f$.*

We say that a dag $D$ on a set $N$ of nodes is *admissible* for $f$ if the skeleton of $D$ is a subgraph of the super-structure $S_f$. Furthermore, we say that a dag $D$ on $N$ is *strictly admissible* for $f$ if for every node $v \in N$ we have $\mathrm{P}_D(v) \in \mathcal{P}_f(v)$. Note that every strictly admissible dag is also admissible. Furthermore, there always exists a (strictly) admissible dag with the highest score as shown by the following lemma.

**Lemma 1.** *Let $f$ be a local score function defined on a set $N$ of nodes and let $D$ be a dag on $N$. Then there is a strictly admissible dag $D'$ on $N$ with the same score as $D$.*

*Proof.* If $D$ is not (strictly) admissible, i.e., if there is a $v \in N$ such that $f(v, \mathrm{P}_D(v)) = 0$, we can delete all arcs $(w, v)$ such that $w \in \mathrm{P}_D(v)$. This does not decrease the score since $f(v, \emptyset) \geq f(v, \mathrm{P}_D(v)) = 0$ for every such a $v$. $\qquad\square$

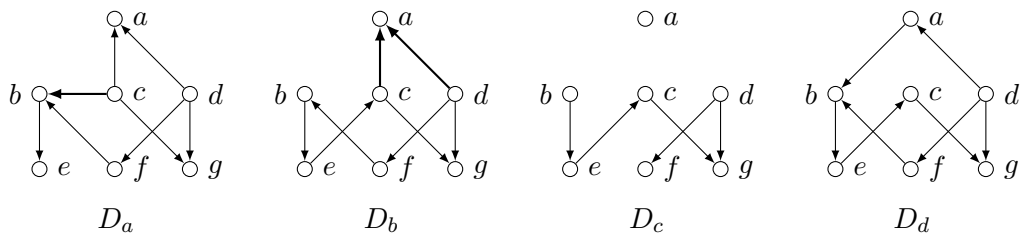

Figure 3: Various dags in correspondence to the local-score function $f$ of Example 1.

**Example 2.** *Figure 3 shows four examples of dags on the the nodes $a, b, c, d, e, f, g$. Using the local-score function $f$ as defined in Example 1, we make the following observations:*

a) *$D_a$ is not admissible, because the super-structure $S_f$ does not contain the edge $\{c, b\}$. The score for $D_a$ is $f(D_a) = 4$.*

b) *$D_b$ is admissible but not strictly admissible, because the node $a$ has parents $c$ and $d$ but $f(a, \{c, d\}) = 0$. The score for $D_b$ is $f(D_b) = 5$.*

c) *$D_c$ is the strictly admissible dag obtained from $D_b$ as described in the proof of Lemma 1. We have $f(D_c) = f(D_b) = 5$.*

d) *$D_d$ is strictly admissible and is also an optimal dag for $f$. The score for $D_d$ is $f(D_d) = 7$.*

### 3.1 Tree Decompositions

When presenting algorithms for graphs of bounded treewidth it is convenient to consider tree decompositions in the following normal form (Kloks, 1994): A triple $(T, \chi, r)$ is a *nice tree decomposition* of a graph $G$ if $(T, \chi)$ is a tree decomposition of $G$, the tree $T$ is rooted at node $r$, and each node of $T$ is of one of the following four types:

1. a *leaf node*: a node having no children;

2. a *join node*: a node $t$ having exactly two children $t_1, t_2$, and $\chi(t) = \chi(t_1) = \chi(t_2)$;

3. an *introduce node*: a node $t$ having exactly one child $t'$, and $\chi(t) = \chi(t') \cup \{v\}$ for a node $v$ of $G$;

4. a *forget node*: a node $t$ having exactly one child $t'$, and $\chi(t) = \chi(t') \setminus \{v\}$ for a node $v$ of $G$.

For convenience we will also assume that $\chi(r) = \emptyset$ for the root $r$ of $T$. This can always be achieved by adding forget nodes on top of the root (see Figure 4 for an example). For a nice tree decomposition $(T, \chi, r)$ we define $\chi^*(t)$ to be the union of all the sets $\chi(t')$ where $t'$ is contained in the subtree of $T$ rooted at $t$. Furthermore, we denote by $F_t$ the set of nodes that have already been "forgotten" at node $t$, i.e.,

$$F_t = \chi^*(t) \setminus \chi(t).$$

As stated in Section 2.3 one of the main properties of tree decompositions that allows for efficient algorithms is the well-known separator property made precise in Proposition 2. The following propositions provide different versions of this separator property of tree decompositions for each of the node types of a nice tree decomposition. Because these propositions are well-known (Kloks, 1994) and are immediate consequences of the separator property of tree decompositions we state them without proofs. The propositions summarize the algorithmic properties of nice tree decompositions that we will use for the design of our algorithm in Section 4.

**Proposition 3.** *Let $t$ be a join node with children $t_1$ and $t_2$. Then $F_{t_1} \cap F_{t_2} = \emptyset$ and there is no edge between a node $u \in F_{t_1}$ and a node $v \in F_{t_2}$ in $G$.*

**Proposition 4.** *Let $t$ be an introduce node with child $t'$ such that $\chi(t) = \chi(t') \cup \{v_0\}$. Then there is no edge from $v_0$ to a node $v \in F_t$ in $G$. Furthermore, $v_0 \notin F_{t'}$.*

**Proposition 5.** *Let $t$ be a forget node with child $t'$ such that $\chi(t) = \chi(t') \setminus \{v_0\}$. Then there is no edge from $v_0$ to a node $v \in V(G) \setminus \chi^*(t)$ in $G$.*

Given a tree decomposition of a graph $G$ of width $w$, one can effectively obtain in time $O(|V(G)|)$ a nice tree decomposition of $G$ with $O(|V(G)|)$ nodes and of width at most $w$ (Kloks, 1994).

**Example 3.** *Figure 4 shows a tree decomposition and a corresponding nice tree decomposition of the super-structure of Example 1.*

## 4. A Dynamic Programming Algorithm for Exact Bayesian Network Structure Learning

In this section we present the dynamic programming algorithm and establish our tractability results. For the remainder of this section $w$ denotes an arbitrary but fixed constant. Recall from the previous section that $\delta_f$ is the maximum number of potential parent sets of a node.

**Theorem 1.** *Given a set $N$ of nodes and a local score function $f$ on $N$ whose super-structure $S_f = (N, E_f)$ has treewidth bounded by an arbitrary constant $w$. Then we can find in time $O(\delta_f^{2(w+1)} \cdot |N|)$ a dag $D$ on $N$ with maximal score $f(D)$.*

Before we devise an algorithm to show Theorem 1 we state and prove a direct consequence of the theorem.

**Corollary 1.** EXACT BAYESIAN NETWORK STRUCTURE LEARNING *can be decided in polynomial time for instances where the super-structure has bounded treewidth. The problem can be decided in linear time if additionally the super-structure has bounded maximum degree.*

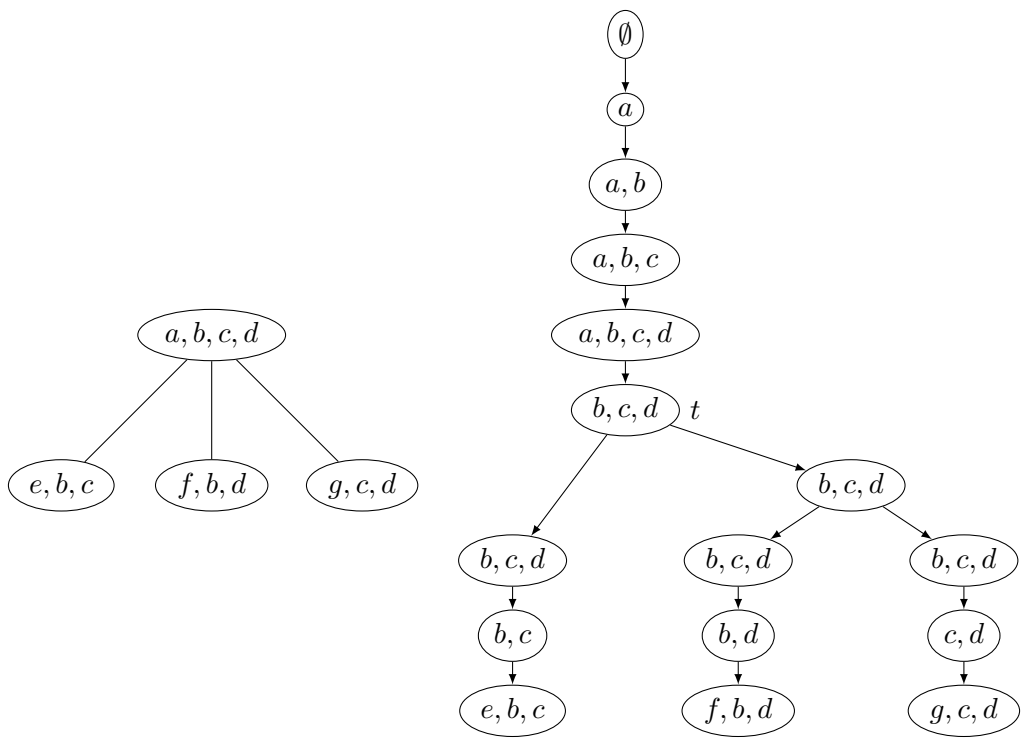

Figure 4: A tree decomposition (left) and a corresponding nice tree decomposition (right) for the super-structure of Example 1.

*Proof.* The first statement follows immediately from the theorem since $\delta_f$ is bounded by the total input size of the instance and $w$ is a constant. Recall from Section 3 that the local score function $f$ is given as the list of all tuples for which $f$ is non-zero and hence $\delta_f$ is bounded by the total input size of the instance. The second statement follows since $\delta_f$ is bounded whenever the maximum degree $d$ of the super-structure is bounded as clearly $\delta_f \leq 2^d$. □

In the following we will assume that we are given a set of nodes $N$ and a local score function $f$ on $N$ together with a nice tree decomposition $(T, \chi, r)$ for $S_f$ of width at most $w$. We are going to establish Theorem 1 by means of a dynamic programming algorithm along a nice tree decomposition for $S_f$, computing local information at the nodes of the tree decomposition that can then be put together to form an optimal dag. Our algorithm closely follows the general approach used by algorithms on graphs (or structures) of bounded treewidth (Bodlaender & Koster, 2008).

A *partial solution* for a tree node $t \in V(T)$ is a dag that can be obtained as the induced subdigraph $D[\chi^*(t)]$ of a strictly admissible dag $D$ for $f$. For a tree node $t$ let $\mathcal{D}(t)$ denote the set of all partial solutions for $t$. For a partial solution $D \in \mathcal{D}(t)$ we set

$$f_t(D) = \sum_{v \in F_t} f(v, \mathrm{P}_D(v)),$$

i.e., $f_t(D)$ is the sum of the scores of the nodes in $F_t$. Recall from the previous section that $F_t$ is the set of forgotten nodes at $t$.

The main idea underlying our algorithm is to reduce the space required to store a partial solution with the help of a so-called record. This becomes possible because of the properties of a tree decomposition manifested by Propositions 3, 4 and 5.

A *record* of a tree node $t \in V(T)$ is a triple $R = (a, p, s)$ such that:

1. $a$ is a mapping $\chi(t) \to \mathcal{P}_f(v)$, i.e., for every $v \in \chi(t)$ we have $a(v) \in \mathcal{P}_f(v)$;

2. $p$ is a transitive binary relation on $\chi(t)$;

3. $s$ is a non-negative real number.

Informally, for a tree node $t \in V(T)$ and a record $R = (a, p, s)$, the mapping $a$ fixes the parent set of every node in $\chi(t)$, $p$ is a compact representation of the reachability relation between the nodes in $\chi(t)$ (using directed paths between nodes in $\chi^*(t)$), and $s$ is the sum of the scores of the nodes that have been forgotten for $t$, i.e., the nodes in $F_t$.

We say that a record *represents* a partial solution $D \in \mathcal{D}(t)$ if it satisfies the following conditions:

1. $a(v) \cap V(D) = \mathrm{P}_D(v)$ for every $v \in \chi(t)$.

2. For every pair of nodes $v_1, v_2 \in \chi(t)$ it holds that $(v_1, v_2) \in p$ if and only if $D$ contains a directed path from $v_1$ to $v_2$.

We say that a record $R = (a, p, s)$ of a tree node $t \in V(T)$ is *valid* if it represents some dag $D \in \mathcal{D}(t)$ and $s$ is the maximum score $f_t(D)$ over all dags in $\mathcal{D}(t)$ represented by $R$. We say a partial solution $D$ that is represented by $R$ is *maximal* with respect to $R$ if $R$ is valid and $f_t(D) = s$. With each tree node $t \in V(T)$ we associate the set $\mathcal{R}(t)$ of all valid records representing partial solutions in $\mathcal{D}(t)$.

In a certain sense, $\mathcal{R}(t)$ is a succinct representation of the optimal elements of $\mathcal{D}(t)$, using space that only depends on $w$ and $\delta_f$, but not on $|N|$.

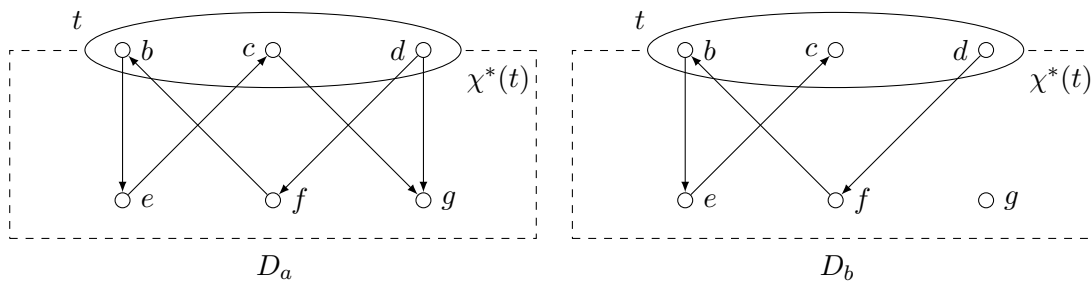

Figure 5: Two partial solutions for the node $t$ of the nice tree decomposition from Example 3.

**Example 4.** *Figure 5 shows two partial solutions $D_a$ and $D_b$ for the node $t$ of the nice tree decomposition Example 3. $D_a$ and $D_b$ are represented by all records $R = (a, p, s)$ with* $a(b) = \{a, f\}$, $a(c) = \{e\}$, $a(d) = \emptyset$ and $p = \{(b, c), (d, b), (d, c)\}$. We have $f_t(D_a) = 3 > f_t(D_b) = 2$ and it is easy to see that $3$ is the maximum score over all partial solutions represented by $R$. Hence the record $R = (a, p, 3)$ is a valid record for $t$ and $D_a$ is maximal with respect to $R$, thus $R \in \mathcal{R}(t)$.

Our dynamic programming algorithm computes the set of all valid records in a bottom up manner, i.e., starting from the leave nodes of the nice tree decomposition the algorithm proceeds to the root node. The next three lemmas show how to compute the set of all valid records for the introduce, forget and join nodes of the nice tree decomposition from the valid records of its children. Informally, if $t$ is an introduce node with child $t'$ such that $\chi(t) = \chi(t') \cup \{v_0\}$ then we compute the set $\mathcal{R}(t)$ of all valid records for $t$ by checking for each potential parent set $P \in \mathcal{P}(v_0)$ for $v_0$ and each valid record $R \in \mathcal{R}(t')$ for $t'$ whether the combination of $P$ and $R$ constitutes a valid record for $t$.

**Lemma 2** (introduce node). *Let $t$ be an introduce node with child $t'$. Then $\mathcal{R}(t)$ can be computed from $\mathcal{R}(t')$ in time $O(\delta_f^{w+1})$.*

*Proof.* In the following we denote by $v_0$ the node introduced by $t$, i.e., $\chi(t) = \chi(t') \cup \{v_0\}$. We are going to establish the lemma with the help of the following claim whose proof can be found in the appendix.

**Claim 1.** *$\mathcal{R}(t)$ is the set of all records $R = (a, p, s)$ such that there is a set $P \in \mathcal{P}_f(v_0)$ and a record $R' = (a', p', s') \in \mathcal{R}(t')$ with:*

1. *$a(v_0) = P$.*

2. *For every $v \in \chi(t')$ it holds that $a(v) = a'(v)$.*

3. *$s = s'$.*

4. *$p$ is the transitive closure of the relation $p' \cup \{ (u, v_0) : u \in P \} \cup \{ (v_0, u) : u \in \chi(t')$ such that $v_0 \in a'(u) \}$.*

5. *$p$ is irreflexive.*

It follows that $\mathcal{R}(t)$ can be computed by checking for every pair $(P, R')$, with $P \in \mathcal{P}_f(v_0)$ and $R' \in \mathcal{R}(t')$, whether it satisfies the conditions (1)–(5). Since there are at most $\delta_f$ possible sets $P$ and at most $O(\delta_f^w)$ possible valid records for $t'$ (observe that $|\chi(t')| \leq w$) the lemma follows from the fact that for every pair $(P, R')$ the conditions can be checked in time that only depends on $w$. $\square$

Informally, if $t$ is a forget node with child $t'$ such that $\chi(t) = \chi(t') \setminus \{v_0\}$ then we compute the set $\mathcal{R}(t)$ of all valid records for $t$ by "projecting" the set $\mathcal{R}(t')$ of valid records for $t'$ to $\chi(t)$.

**Lemma 3** (forget node). *Let $t$ be a forget node with child $t'$. Then $\mathcal{R}(t)$ can be computed from $\mathcal{R}(t')$ in time $O(\delta_f^{w+1})$.*

*Proof.* In the following we denote by $v_0$ the forgotten node, i.e. $\chi(t) = \chi(t') \setminus \{v_0\}$. We will show that $\mathcal{R}(t)$ can be obtained as the "projection" of $\mathcal{R}(t')$ to $\chi(t)$. Before doing so we need some additional notation. Let $R' = (a', p', s') \in \mathcal{R}(t')$. We define the *projection* of $R'$ to $t$, denoted $R'[t]$ as the record $R = (a, p, s)$ such that:

1. $a$ is the restriction of $a'$ to $\chi(t)$.

2. $p = \{ (v, w) \in p' : v, w \in \chi(t) \}$.

3. $s = s' + f(v_0, a'(v_0))$.

Furthermore, we define the projection of $\mathcal{R}(t')$ to $t$, denoted $\mathcal{R}(t')[t]$, as the set of all records $R'[t]$ for $R' \in \mathcal{R}(t')$. We say that a record $R = (a, p, s) \in \mathcal{R}(t')[t]$ is maximal if there is no $s'$ with $s' > s$ such that $(a, p, s') \in \mathcal{R}(t')[t]$.

Again, we are going to establish the lemma with the help of the following claim whose proof can be found in the appendix.

**Claim 2.** $\mathcal{R}(t)$ *is the set of all maximal records in* $\mathcal{R}(t')[t]$.

Since $\mathcal{R}(t')$ contains at most $O(\delta_f^{w+1})$ records it is easy to see that $\mathcal{R}(t)$ can be computed in time $O(\delta_f^{w+1})$. □

Informally, if $t$ is a join node with children $t_1$ and $t_2$ then we compute the set $\mathcal{R}(t)$ of all valid records for $t$ by checking for each record $R_1 \in \mathcal{R}(t_1)$ and each record $R_2 \in \mathcal{R}(t_2)$ whether the combination of $R_1$ and $R_2$ constitutes a valid record for $t$.

**Lemma 4** (join node)**.** *Let* $t_1$, $t_2$ *be the children of* $t$ *in* $T$. *Then* $\mathcal{R}(t)$ *can be computed from* $\mathcal{R}(t_1)$ *and* $\mathcal{R}(t_2)$ *in time* $O(\delta_f^{2(w+1)})$.

*Proof.* We are going to establish the lemma with the help of the following claim (the rather technical proof of this claim can be found in the appendix).

**Claim 3.** $\mathcal{R}(t)$ *is the set of all records* $R = (a, p, s)$ *such that there are records* $R_1 = (a_1, p_1, s_1) \in \mathcal{R}(t_1)$ *and* $R_2 = (a_2, p_2, s_2) \in \mathcal{R}(t_2)$ *with:*

1. $a = a_1 = a_2$.

2. $s = s_1 + s_2$.

3. $p$ *is the transitive closure of* $p_1 \cup p_2$.

4. $p$ *is irreflexive, i.e., there is no* $v \in \chi(t)$ *such that* $(v, v) \in p$.

It follows that $\mathcal{R}(t)$ can be computed by considering all pairs of records $R_1 \in \mathcal{R}(t_1)$ and $R_2 \in \mathcal{R}(t_2)$ and checking conditions (1)–(4). Since there are at most $O(\delta_f^{w+1})$ valid records for every $t \in V(T)$ and for every such pair of records the time required to check the conditions (1)–(4) does only depend on $w$, it follows that the running time of this procedure is at most $O(\delta_f^{2(w+1)})$. □

We are now ready to establish Theorem 1.

*Proof of Theorem 1.* Let $N$ be a set of nodes, $f$ a local score function on $N$ where the super-structure $S_f$ has treewidth $w$ (a constant) and $|N| = n$. We compute a nice tree decomposition $(T, \chi, r)$ of $S_f$ of width $w$ and with $O(n)$ nodes. This can be accomplished in time $O(n)$ (see the discussion in Section 2.3).

Next we compute the sets $\mathcal{R}(t)$ via a bottom-up traversal of $T$. For a leaf node $t$ we can compute $\mathcal{R}(t)$ just by considering all strictly admissible dags on the at most $w + 1$ nodes in $\chi(t)$. This can clearly be done in time $O(\delta_f^{w+1})$ and we can then use Lemmas 4, 2 and 3 to compute the sets $\mathcal{R}(t)$ for all other $O(n)$ tree nodes in time $O(\delta_f^{2(w+1)} \cdot n)$.

Since $\chi(r) = \emptyset$, the partial solutions for the root $r$ of $T$ are exactly the strictly admissible dags for $f$, and we have $f_r(D) = f(D)$ for each such dag $D$. After the computation of the sets $\mathcal{R}(t)$ for all tree nodes $t$, the set $\mathcal{R}(r)$ contains exactly one record $R = (\emptyset, \emptyset, s)$. By the above considerations, it follows that $s$ is the largest score of all strictly admissible dags for $f$, and, as noted in Section 3, this is also the largest score of any dag whose nodes belong to $N$. It is now easy to compute a dag $D$ with score $f(D) = s$ via a top-down traversal of $T$ starting from $r$ and using the information previously stored at each node in $T$. This can also be accomplished in time $O(\delta_f^{2(w+1)} \cdot n)$. □

We close this section with a remark concerning the relationship between the treewidth of a Bayesian network and the treewidth of its super-structure. For Bayesian reasoning one usually associates with the dag $D$ of a Bayesian network its moral graph $M(D)$ which is the skeleton of $D$ plus edges joining nodes that have a common child in $D$. The treewidth of a Bayesian network is the treewidth of its moral graph (Darwiche, 2001; Dechter, 1999). We observe that for every Bayesian network of bounded treewidth there is a super-structure of bounded treewidth that contains the Bayesian network. Hence, such a Bayesian network can be learned considering only super-structures of bounded treewidth. On the other hand if a Bayesian network is contained in a super-structure of bounded treewidth then the Bayesian network has bounded treewidth under the reasonable assumption that each of its nodes has a bounded number of parents. Consequently, if a Bayesian network is learned from a super-structure of bounded treewidth it is reasonable to assume that the Bayesian network has bounded treewidth as well.

## 5. Refined Complexity Analysis

When presenting our algorithm in Section 4 we focused on a broad evaluation of its complexity. In this Section we provide a more fine-grained analysis of its running time.

In the following we assume that $N$ is a set of nodes, $f$ is a local score function on $N$, and $(T, \chi, r)$ is a nice tree decomposition of $S_f$ of width at most $w$. We can improve on the running time of our algorithm by using the following five ideas.

I1 By keeping the records for each node in the tree decomposition in some fixed order we can improve the time needed at a join node of the tree decomposition from $O(\delta_f^{2(w+1)})$ to $O(\delta_f^{w+1})$ without any additional cost for sorting. We achieve this by generating new records in an ordered manner and keeping track of that order when the records are stored.

I2 Parent sets that are supersets of parent sets with a higher score can be disregarded (de Campos, Zeng, & Ji, 2009). This simple preprocessing rule usually allows us to consider fewer than the potentially $2^d$ potential parent sets. In the sequel we will denote the preprocessed score function by $f'$. Clearly, $|\mathcal{P}_f(v)| \leq |\mathcal{P}_{f'}(v)|$ for every $v \in N$.

I3 If the maximum in-degree of the resulting BN is bounded in advance one can use this fact to further preprocess the score function in the natural way. In the following we denote the resulting score function by $f''$. Clearly, $|\mathcal{P}_{f''}(v)| \leq |\mathcal{P}_{f'}(v)|$ for every $v \in N$.

I4 Every node of the network might have a different number of potential parent sets. Consequently, instead of using one upper bound for the number $\delta_{f''}$ of potential parent sets it is more realistic to consider the actual number $|\mathcal{P}_{f''}(v)|$ of potential parent sets for each node $v \in N$.

I5 Only valid records need to be stored by our algorithm, i.e., records that represent acyclic networks.

Considering the ideas I1–I4 the worst-case complexity of our algorithm can be refined as follows:

$$\sum_{t \in V(T)} O(w^2 \cdot \prod_{v \in \chi(t)} |\mathcal{P}_{f''}(v)|)$$

Observe that $w^2 \cdot \prod_{v \in \chi(t)} |\mathcal{P}_{f''}(v)|$ is the number of potential records for the tree node $t$. Because of idea I5 in general not all of these records need to be stored by our algorithm. This refined analysis suggests that the running time of our algorithm is dominated by the maximum number of records that need to be stored for a tree node $t \in V(T)$.

## 6. Hardness Results for Exact Bayesian Network Structure Learning

The following result follows by a reduction due to Chickering (1996).

**Theorem 2.** EXACT BAYESIAN NETWORK STRUCTURE LEARNING *is* NP-*hard for instances with super-structures of maximum degree* 4.

*Proof.* Since we use a well-known reduction from Chickering, we will only sketch the argument. The reduction is from FEEDBACK ARC SET (FAS). The problem asks whether a digraph $D = (V, E)$ can be made acyclic by deleting at most $k$ arcs (the deleted arcs form a feedback arc set of $D$). The problem is NP-hard for digraphs with skeletons of maximum degree 4 (Karp, 1972). Given an instance $(D, k)$ of FAS, where the skeleton of $D$ has maximum degree 4, we construct a set $V' = V(D) \cup E(D)$ of nodes and a local score function $f$ on $V'$ by setting $f((u, v), \{u\}) = 1$ for all $(u, v) \in E(D)$, $f(v, \{(u, v) : u \in \mathrm{P}_D(v)\}) = |\mathrm{P}_D(v)|$ for all $v \in V(D)$, and $f(v, P) = 0$ in all other cases. Clearly, the super-structure $S_f$ (recall the definition of $S_f$ in Section 3) is isomorphic to the undirected graph obtained from the skeleton of $D$ after subdividing every edge once, hence the maximum degree of $S_f$ is at most the maximum degree of $D$ which is 4. It is easy to see that $D$ has a feedback arc set of size $\leq k$ if and only if there exists a dag $D'$ whose skeleton is a spanning subgraph of $S_f$ with $f(D') \geq 2 \cdot |E| - k$. □

**Theorem 3.** Exact Bayesian Network Structure Learning *parameterized by the treewidth of the super-structure is* W[1]*-hard.*

*Proof.* We devise an fpt-reduction from the following problem, which is well-known to be W[1]-complete (Pietrzak, 2003).

| | |
|---|---|
| Partitioned Clique | |
| Input: | A $k$-partite graph $G = (V, E)$ with partition $V_1, \ldots, V_k$ such that $|V_i| = |V_j| = n$ for $1 \leq i < j \leq k$. |
| Parameter: | The integer $k$. |
| Question: | Are there nodes $v_1, \ldots, v_k$ such that $v_i \in V_i$ for $1 \leq i \leq k$ and $\{v_i, v_j\} \in E$ for $1 \leq i < j \leq k$? (The graph $K = (\{v_1, \ldots, v_k\}, \{\{v_i, v_j\} : 1 \leq i < j \leq k\})$ is a $k$-*clique* of $G$.) |

Let $G = (V, E)$ be an instance of this problem with partition $V_1, \ldots, V_k$, $|V_1| = \cdots = |V_k| = n$ and parameter $k$. Informally, we encode the given instance of Partitioned Clique as an instance $(N, f, s)$ of Exact Bayesian Structure Learning such that $G$ has a $k$-clique if and only if there is a Bayesian network $D$ with $f(D) \geq s$. To achieve this we introduce a node $n_v$ for every node $v$ of $G$ and one node $a_{ij}$ for every $1 \leq i \neq j \leq k$. Then a node that corresponds to a node in $V_i$ achieves its maximum score for the parent set that contains all nodes $a_{ij}$ for $1 \leq j \leq k$. A node $a_{ij}$ achieves its maximum score if its parent set corresponds to an edge of $G$ between a node in $V_i$ and a node in $V_j$. Hence, the edges of $G$ are encoded in the local score function of the nodes $a_{ij}$ for $1 \leq i \neq j \leq k$. By choosing the proper scores for these nodes and the right threshold $s$ we can assure that in any Bayesian network whose score is higher than $s$ every node $a_{ij}$ has to attain its maximum score, and all but at most $k$ nodes that correspond to nodes of $G$ do not achieve their maximum score. It follows that the parent sets chosen for the nodes $a_{ij}$ correspond to the edges of a $k$-clique of $G$.

In order to make later calculations easier we will assume that $k > 2$ for the remainder of this proof. Let $\alpha = k^2 - k - 1$, $\varepsilon = 2k$ and $s = nk\alpha + 1$. We construct a set $N$ of nodes and a local score function $f$ on $N$ satisfying the following claims.

**Claim 4.** $\operatorname{tw}(S_f) \leq k(k-1)/2$

**Claim 5.** $G$ *has a $k$-clique if and only if there is a dag $D$ such that $f(D) \geq s$.*

We will have shown the theorem after establishing the two claims.

We set $A = \{a_{ij} : 1 \leq i < j \leq k\}$, $N = V(G) \cup A$, and $A_i = \{a_{lk} \in A : l = i \text{ or } k = i\}$ for every $1 \leq i \leq k$. We are now ready to define $f$. We set $f(v, A_i) = \alpha$ for every $v \in V_i$, and $f(a_{ij}, \{u, w\}) = \varepsilon$ for every $1 \leq i < j \leq k$, $u \in V_i$, $w \in V_j$, and $\{u, w\} \in E(G)$. Furthermore, we set $f(v, P) = 0$ for all the remaining combinations of $v$ and $P$. See Figure 6 for an illustration. Now, Claim 4 follows from Proposition 1 with $X = A$. Hence, it remains to show Claim 5.

Before we go on to show Claim 5 we give some further notation and explanation. Let $E' \subseteq E(G)$. We denote by $V(E')$ the set of all nodes incident to edges in $E'$, i.e., the set $\bigcup_{e \in E'} e$. We say that $E'$ is *representable* if for every $1 \leq i < j \leq k$ it contains at most one edge between a node in $V_i$ and a node in $V_j$. We define $e_{ij}(E') = \{v_i, v_j\}$ if $E'$ contains the edge between $v_i \in V_i$ and $v_j \in V_j$ and $e_{ij} = \emptyset$ otherwise. We define $D(E')$ to be the directed graph with node set $N$ and arc set $\{(v, a_{ij}) : v \in e_{ij}(E') \text{ and } 1 \le i < j \le k\} \cup \{(a_{ij}, v) : v \notin e_{ij}(E') \text{ and } 1 \le i < j \le k\}$. Figure 7 shows $D(E')$ for a representable edge set of the example graph $G$ from Figure 6.

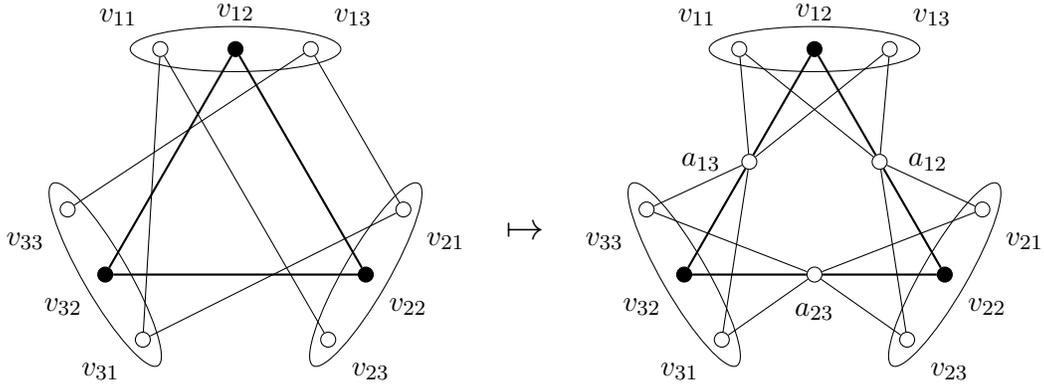

Figure 6: An example for the graph $G$ and the super-structure $S_f$ according to the construction used in the proof of Theorem 3.

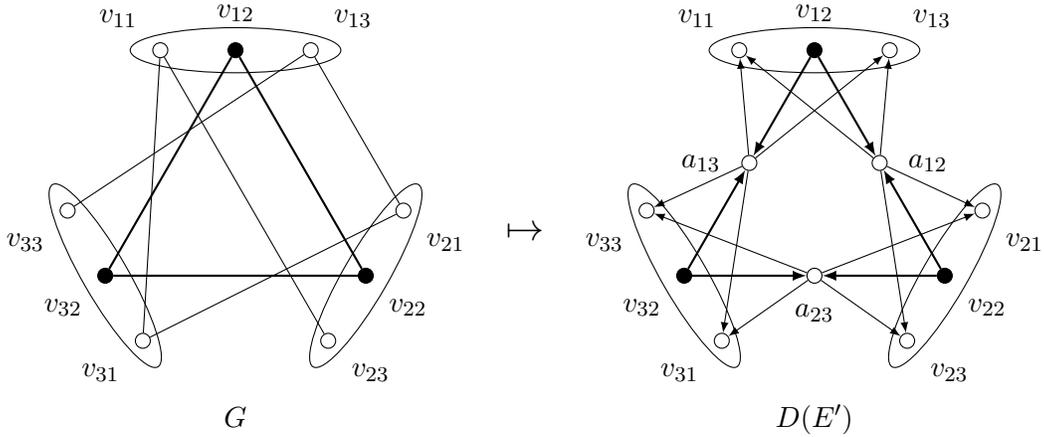

Figure 7: The example graph $G$ from Figure 6 together with an edge set $E'$, given as the bold edges in the illustration of $G$, and the resulting dag $D(E')$.

The main idea to show Claim 5 is that $f(D) \ge s$ if and only if $D$ has the form $D(E')$ for a representable edge set $E'$ that corresponds to a $k$-clique in $G$. This is formally expressed by the following claim whose proof can be found in the appendix.

**Claim 6.** *The following statements are equivalent:*

1. *$G$ has a $k$-clique.*

2. *There is a dag $D$ with $f(D) \ge s$.*

3. *There is a representable edge set $E' \subseteq E(G)$ such that $f(D(E')) \ge s$.*

□

Note that in contrast to Theorem 2, it is essential for Theorem 3 that the super-structure has unbounded degree: if both degree and treewidth are bounded then the problem is fixed-parameter tractable by Corollary 1 and so unlikely to be W[1]-hard.

## 7. $k$-Neighborhood Local Search

Important and widely used algorithms for BNSL are based on local search methods (Heckerman, Geiger, & Chickering, 1995). Usually the local search algorithm tries to improve the score of a given dag by transforming it into a new dag by adding, deleting, or reversing one arc at a time (in symbols: ADD, DEL, and REV, respectively). The main obstacle for local search methods is the danger of getting stuck at a poor local optimum. A possibility for decreasing this danger is to perform $k > 1$ elementary changes in one step, known as $k$-*Neighborhood Local Search* or $k$-*Local Search* for short. For BNSL, when we try to improve the score of a dag on $n$ nodes, the $k$-local search space is of order $n^{O(k)}$. Therefore, if carried out by brute-force, $k$-local search is too costly even for small values of $k$. It is therefore not surprising that most practical local search algorithms for BNSL consider 1-neighborhoods only.

The study of the parameterized complexity of $k$-local search was initiated by Fellows (2003). To date a collection of positive and negative results on the parameterized complexity of $k$-local search for various combinatorial optimization problems are known. For instance, $k$-local search has already been investigated for combinatorial problems on graphs (Khuller, Bhatia, & Pless, 2003; Marx, 2008; Fellows, Rosamond, Fomin, Lokshtanov, Saurabh, & Villanger, 2009; Gaspers, Kim, Ordyniak, Saurabh, & Szeider, 2012), for the problem of finding a minimum weight assignment for a Boolean constraint satisfaction instance (Krokhin & Marx, 2012), for the stable marriage problem with ties (Marx & Schlotter, 2011), and for the satisfiability problem (Szeider, 2011).

In this section we show that $k$-local search for BNSL can be solved in linear time if the super-structure has bounded treewidth and bounded maximum degree. This result is in good agreement with Theorem 1. However, in contrast to Exact BNSL it might still be possible to drop one of these restrictions without losing uniform polynomial-time tractability, but we show that this is not the case. We also investigate $k$-Local Search BNSL for different combinations of allowed operations such as reversal, addition and deletion of an arc. Our results are mostly negative. In fact, somewhat surprisingly, $k$-Local Search BNSL remains hard even if edge reversal is the only allowed operation.

Before we state and show our results we define $k$-Local Search BNSL more formally. Let $k \geq 0$ and $\mathbb{O} \subseteq \{\text{ADD}, \text{DEL}, \text{REV}\}$. Consider a dag $D = (V, E)$. A directed graph $D' = (V', E')$ is a $k$-$\mathbb{O}$-*neighbor* of $D$ if

1. $D'$ is a dag,

2. $V = V'$,

3. $E'$ can be obtained from $E$ by performing at most $k$ operations from the set $\mathbb{O}$.

For $\mathbb{O} \subseteq \{\text{ADD}, \text{DEL}, \text{REV}\}$ we consider the following parameterized decision problem.

| $k$-$\mathbb{O}$-LOCAL SEARCH BAYESIAN NETWORK STRUCTURE LEARNING | |
|---|---|
| Input: | A local score function $f$, a dag $D$ that is admissible for $f$, and an integer $k$. |
| Question: | Is there a $k$-$\mathbb{O}$-neighbor $D'$ of $D$ with a higher score than $D$? |

Note that the problem does not change if we require $D'$ to be admissible, as we can always avoid the addition of an inadmissible arc.

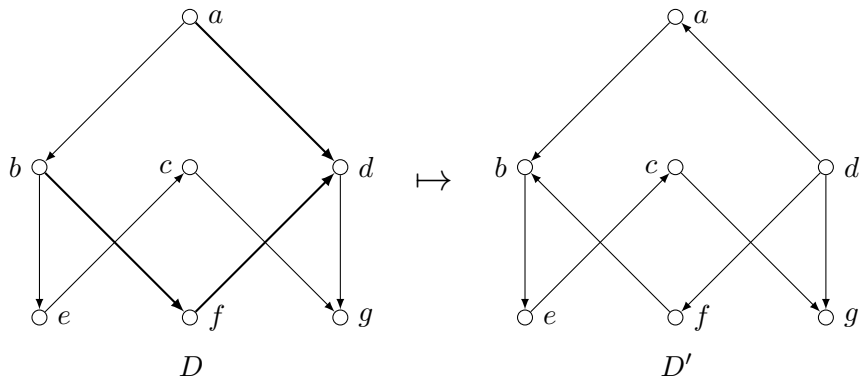

Figure 8: Two dags illustrating the notion of a $k$-$\mathbb{O}$-neighbor in Example 5.

**Example 5.** *Figure 8 shows two dags $D$ and $D'$ such that $D'$ can be obtained from $D$ by either reversing or by deleting and adding the reverse of the bold arcs in $D$. It follows that $D'$ is a 3-{REV}-neighbor and a 6-{ADD, DEL}-neighbor of $D$. The directed graph obtained from $D$ after only reversing the arc $(a, d)$ contains a cycle (on the nodes $\{a, b, f, d\}$) but $D'$ does not. The score of $D'$ is larger than the score of $D$, since $f(D) = 3$ and $f(D') = 7$ (using the score function $f$ as depicted in Figure 2).*

**Proposition 6.** $k$-LOCAL SEARCH BAYESIAN NETWORK STRUCTURE LEARNING *can be decided in linear time for instances where the super-structure has bounded treewidth and bounded maximum degree.*

*Proof.* As the proof uses the same arguments as the proof of Theorem 1 we only sketch the proof of this proposition.

In the following we will assume that we are given an instance $I = (D, f, k)$ of $k$-$\mathbb{O}$-LOCAL SEARCH BAYESIAN NETWORK STRUCTURE LEARNING together with a nice tree decomposition $(T, \chi, r)$ for $S_f$ of width at most $w$ and let $d$ be the maximum degree of $S_f$. As before we assume that $w$ and $d$ are constants. For a set $S$ we denote by $[S]$ the set of all subsets of $S$.

The main difference to the proof of Theorem 1 is that we are now only interested in solutions which are $k$-$\mathbb{O}$-neighbors of $D$. To take this into account we need to slightly extend the concept of a record for a tree node $t \in V(T)$. We include an integer $c$ to reflect the "cost" of a partial solution that is the smallest number of operations from $\mathbb{O}$ needed to obtain $D$. For technical reasons we also include a mapping $b$ that assigns the set of forgotten children to every node contained in $\chi(t)$. This allows us to compute the value of $c$ for a forget node.

A *record* of a tree node $t \in V(T)$ is a quintuple $R = (a, b, p, c, s)$ such that:

1. $a$ is a mapping $\chi(t) \to \mathcal{P}_f(v)$;

2. $b$ is a mapping $\chi(t) \to [F_t]$;

3. $p$ is a transitive binary relation on $\chi(t)$;

4. $c$ is a non-negative integer;

5. $s$ is a non-negative real number.

We say that a record *represents* a partial solution $D_p \in \mathcal{D}(t)$ if it satisfies the following conditions:

1. $a(v) \cap V(D_p) = \mathrm{P}_{D_p}(v)$ for every $v \in \chi(t)$.

2. $b(v) = \{\, u \in F_t : (v, u) \in E(D_p) \,\}$ for every $v \in \chi(t)$.

3. For every pair of nodes $v_1, v_2 \in \chi(t)$ it holds that $(v_1, v_2) \in p$ if and only if $D_p$ contains a directed path from $v_1$ to $v_2$.

4. $c \leq k$ is the smallest integer such that $D_p[F_t]$ is a $c$-$\mathbb{O}$-neighbor of $D[F_t]$.

We say that a record $R = (a, b, p, c, s)$ of a tree node $t \in V(T)$ is *valid* if it represents some dag $D_p \in \mathcal{D}(t)$ and $s$ is the maximum score $f_t(D)$ over all dags in $\mathcal{D}(t)$ represented by $R$. We say a partial solution $D$ that is represented by $R$ is *maximal* with respect to $R$ if $R$ is valid and $f_t(D) = s$. With each tree node $t \in V(T)$ we associate the set $\mathcal{R}(t)$ of all valid records representing partial solutions in $\mathcal{D}(t)$.

It is now straightforward to adapt the dynamic programming algorithm of Section 4 to the new setting. Observe that there are at most $k + 1$ possible values for $c$. Furthermore, because every considered partial solution $D_p$ is admissible, the number of possible values for $b(v)$ for every $v \in \chi(t)$ is bounded by $2^d$. It follows that the space requirement to store such a record is at most $(k+1) \cdot 2^{d(w+1)}$ times the space requirement needed to store a record as defined in Section 4. Using the same argumentation as in Section 4 this leads to an overall running time of $O((\delta_f^{w+1} \cdot (k+1) \cdot 2^{d(w+1)})^2 \cdot |V(D)|) \leq O((d^{w+1} \cdot (k+1) \cdot 2^{d(w+1)})^2 \cdot |V(D)|)$. Since $w$ and $d$ are constants, this constitutes a linear running time. $\qquad\square$

**Theorem 4.** *If $\mathbb{O} = \{\textsc{add}\}$ or $\mathbb{O} = \{\textsc{del}\}$, then $k$-$\mathbb{O}$-Local Search Bayesian Network Structure Learning is solvable in polynomial time.*

*Proof.* We only consider $\mathbb{O} = \{\textsc{add}\}$ as the proof for $\mathbb{O} = \{\textsc{del}\}$ is analogous. Let $I = (D, f, k)$ be the given instance of $k$-$\{\textsc{add}\}$-Local Search Bayesian Network Structure Learning. Since we are only allowed to add arcs to $D$ every step must leave $D$ acyclic. It follows that there is a $k$-$\{\textsc{add}\}$-neighbor $D'$ of $D$ with $f(D') > f(D)$ if and only if there is a node $v \in V(D)$ such that the addition of at most $k$ incoming arcs increases the score of $v$ and the resulting digraph remains acyclic. Now, for every $P \subseteq V(D) \setminus \{v\}$ one can easily check whether $f(v, P) > f(v, \mathrm{P}_D(v))$ and whether $P$ can be obtained from $\mathrm{P}_D(v)$ via the addition of at most $k$ incoming arcs and whether the resulting digraph is acyclic. $\qquad\square$

In view of Theorem 4 let us define a set $\mathbb{O} \subseteq \{\text{ADD}, \text{DEL}, \text{REV}\}$ to be *non-trivial* if $\mathbb{O} \notin \{\emptyset, \{\text{ADD}\}, \{\text{DEL}\}\}$.

**Theorem 5.** *Let $\mathbb{O} \subseteq \{\text{ADD}, \text{DEL}, \text{REV}\}$ be non-trivial. Then $k$-$\mathbb{O}$-LOCAL SEARCH BAYESIAN NETWORK STRUCTURE LEARNING is W[1]-hard for parameter* $\text{tw}(S_f) + k$.

*Proof.* We slightly modify the reduction given in the proof of Theorem 3. Let $I = (G, k)$ be the given instance of PARTIONED CLIQUE and let $N$, $f$ and $s$ be defined in correspondence to $I$ and the proof of Theorem 3. We will distinguish two cases depending on whether it is allowed to reverse an arc or not, i.e., depending on whether $\text{REV} \in \mathbb{O}$.

For the case that $\text{REV} \in \mathbb{O}$ we claim that $I' = (f, D, k')$ where $D = D(\emptyset)$ and $k' = \binom{k}{2}$ is an instance of $k'$-$\mathbb{O}$-LOCAL SEARCH BAYESIAN NETWORK STRUCTURE LEARNING such that $G$ contains a $k$-clique if and only if there is a $k'$-$\mathbb{O}$-neighbor $D'$ of $D$ with $f(D') > f(D)$. To see this let $K$ be a $k$-clique in $G$. It follows from Claim 6 that there is a representable edge set $E'$ with $f(D(E')) \geq s > f(D) = s - 1$ and since $E'$ is representable it is easy to see that $D(E')$ can be obtained from $D$ by the reversal of at most $k'$ arcs in $D$. Hence $D' = D(E')$ is a $k'$-$\mathbb{O}$-neighbor of $D$ with $f(D') > f(D)$. To see the reverse let $D'$ be a $k'$-$\mathbb{O}$-neighbor of $D$ with $f(D') > f(D) = s - 1$. Hence $D'$ is a dag and since $f(D')$ is integer it follows that $f(D') \geq s$. Again, using Claim 6, we have that there is a $k$-clique in $G$.

Now, for the only remaining case, i.e., the case that $\mathbb{O} = \{\text{ADD}, \text{DEL}\}$ we claim that $I'' = (f, D, k'')$ where $k'' = 2k'$ is an instance of $k'$-$\mathbb{O}$-LOCAL SEARCH BAYESIAN NETWORK STRUCTURE LEARNING such that $G$ contains a $k$-clique if and only if there is a $k''$-$\mathbb{O}$-neighbor $D'$ of $D$ with $f(D') > f(D)$. The proof uses the same arguments as in the case that $\text{REV} \in \mathbb{O}$ only that we now need twice as many operations. That is, we have to replace the reversal of an arc $(u, v)$ with the deletion of the arc $(u, v)$ followed by the addition of the arc $(v, u)$. $\qquad\square$

In the preliminary version of this paper (Ordyniak & Szeider, 2010) we showed the following theorem by a parametrized reduction from RED/BLUE NON-BLOCKER, which had been claimed to be W[1]-complete for graphs of bounded degree (Downey & Fellows, 1995). However, recently we found out that this problem is in fact fixed-parameter tractable and that the proof published in the work of Downey and Fellows (1995) contained a mistake (Fellows, 2012). We therefore use a reduction from INDEPENDENT SET that does not require the original instance to have bounded degree. This even allows us to strengthen our result by decreasing the upper bound on the maximum degree of the super-structure from 5 to 3.

**Theorem 6.** *Let $\mathbb{O} \subseteq \{\text{ADD}, \text{DEL}, \text{REV}\}$ be non-trivial. Then $k$-$\mathbb{O}$-LOCAL SEARCH BAYESIAN NETWORK STRUCTURE LEARNING is W[1]-hard for parameter $k$. Hardness even holds if the super-structure $S_f$ has maximum degree 3.*

*Proof.* We devise an fpt-reduction from the following problem which is known to be W[1]-complete (Downey & Fellows, 1999).

<div style="border:1px solid black">

**Independent Set**

| | |
|---|---|
| Input: | An undirected graph $G = (V, E)$ and an integer $k$. |
| Parameter: | The integer $k$. |
| Question: | Does $G$ have an independent set of size at least $k$, i.e., is there a set $S \subseteq V$ with $|S| \geq k$ such that $\{u, v\} \notin E$ for every pair of nodes $u, v \in S$. |

</div>

To simplify the initial construction we first prove the theorem for the case that the maximum degree of the super-structure is at most 5. We later show how to refine the proof for super-structures with maximum degree at most 3.

Let $(G = (V, E), k)$ be an instance of this problem and $k' = 2k + 1$ if REV $\in \mathbb{O}$ and $k' = 2(2k + 1)$ otherwise. We construct a dag $D$ and a local score function $f$ such that $G$ has an independent set of size at least $k$ if and only if $D$ has a $k'$-$\mathbb{O}$-neighbor $D'$ with a higher score than $D$.

The dag $D$ is obtained from $G$ by applying the following steps (see Figure 9 for an illustration):

1. We replace every node $v \in V$ with the two nodes $v^1$ and $v^2$ and an arc $(v^1, v^2)$.

2. For every node $v \in V$ we add a binary tree $T_{v^1}$ with exactly $|N_G(v)|$ leaves. The root of $T_{v^1}$ is $v^1$ and all arcs of $T_{v^1}$ are directed away from $v^1$. Furthermore, we define $l_{v^1}$ to be a bijective mapping from $N_G(v)$ to the leaves of $T_{v^1}$.

3. For every node $v \in V$ we add a binary tree $T_{v^2}$ with exactly $|N_G(v)|$ leaves. The root of $T_{v^2}$ is $v^2$ and all arcs of $T_{v^2}$ are directed towards $v^2$. Furthermore, we define $l_{v^2}$ to be a bijective mapping from $N_G(v)$ to the leaves of $T_{v^2}$.

4. For every edge $\{u, v\} \in E$, we add the arcs $(l_{u^1}(v), l_{v^2}(u))$ and $(l_{v^1}(u), l_{u^2}(v))$ to $D$.

5. We add a binary tree $T_1$ with root $r_1$ with exactly $|V|$ leaves, whose edges are directed away from $r_1$. We define $l_{T_1}$ to be a bijective mapping from $V$ to the leaves of $T_1$.

6. We add a binary tree $T_2$ with root $r_2$ with exactly $|V|$ leaves, whose edges are directed towards $r_2$. We define $l_{T_2}$ to be a bijective mapping from $V$ to the leaves of $T_2$.

7. For every $v \in V(G)$, we add the arcs $(v^1, l_{T_1}(v))$ and $(v^1, l_{T_2}(v))$ to $D$.

8. We add the arc $(r_2, r_1)$ to $D$.

This completes the construction of $D$. Next we define the local score function $f$ on $V(D)$. Let $\alpha = k - 1$, $\beta = |V(G)|$ and $\varepsilon = 1$.

1. For every $n \in V(D) \setminus \{ v^1 : v \in V(G) \} \cup \{r_1\}$ we set $f(n, \mathrm{P}_D(n)) = \beta$.

2. For every $v \in V(G)$ we set $f(v^1, \{v^2, l_{T_1}(v)\}) = \varepsilon$, $f(v^2, \mathrm{P}_D(v^2) \setminus \{v^1\}) = \beta$, and $f(l_{T_1}(v), \mathrm{P}_D(l_{T_1}(v)) \setminus \{v^1\}) = \beta$.

3. We set $f(r_1, \mathrm{P}_D(r_1)) = \alpha$ and $f(r_2, \mathrm{P}_D(r_2) \cup \{r_1\}) = \beta$.

4. For all the remaining combinations of $n \in V(D)$ and $P \subseteq V(D)$ we set $f(n, P) = 0$.

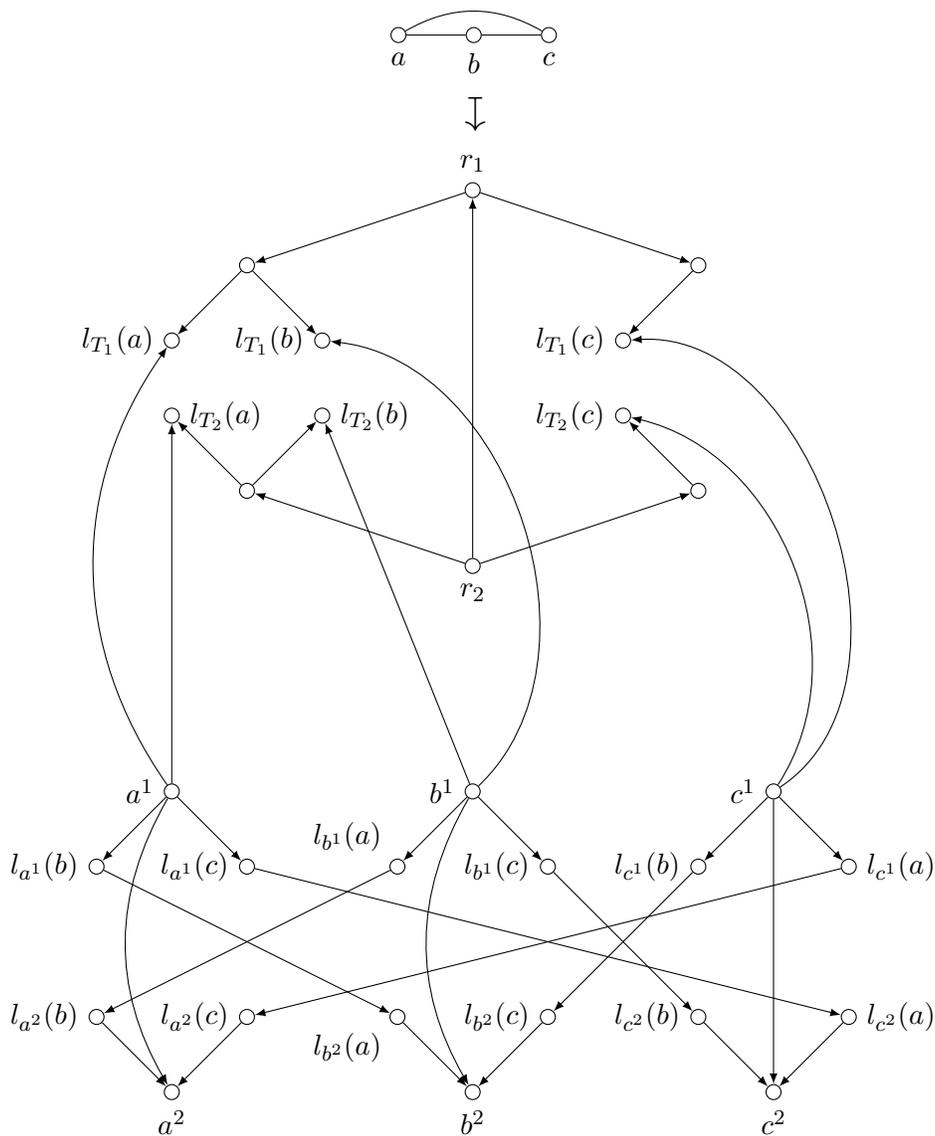

Figure 9: Top: An example graph $G$. Bottom: The dag $D$ resulting from $G$ using the construction in the proof of Theorem 6.

Evidently $D$ is acyclic and both $D$ and $f$ can be constructed from $G$ in polynomial time. Observe that the super-structure $S_f$ is exactly the skeleton of $D$. Hence, by construction, the nodes $v^1$ for $v \in V(D)$ have degree at most 5 while all other nodes of $S_f$ have degree at most 3 showing that the maximum degree of $S_f$ is at most 5. Consequently, we can establish the theorem with the help of the following claim whose proof can be found in the appendix.

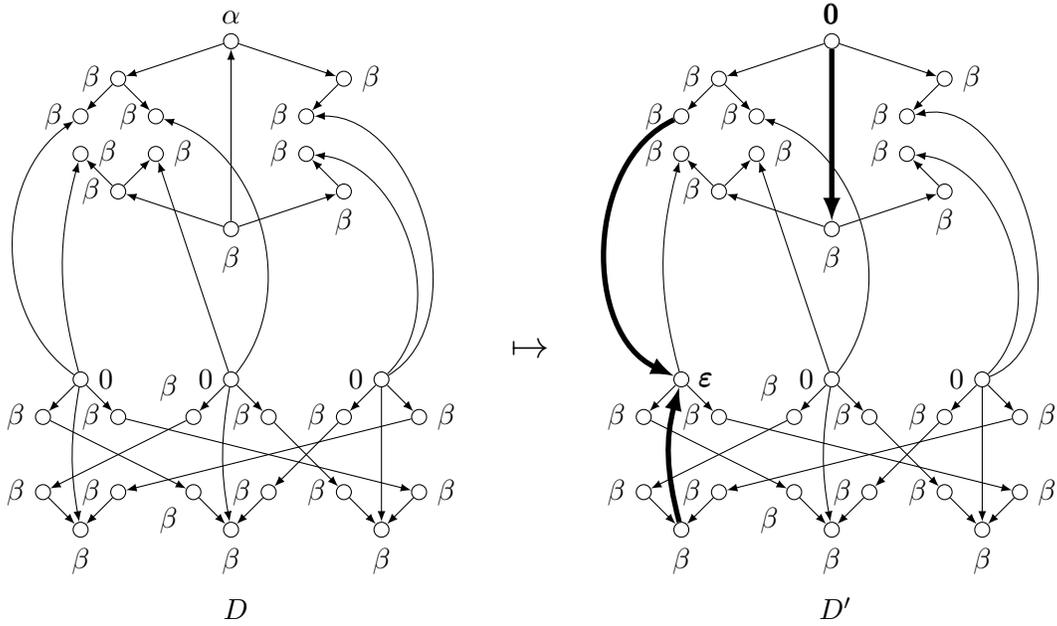

Figure 10: The dag $D$ from Figure 9 together with a $k'$-{Rev}-neighbor $D'$ with $f(D') > f(D)$. Here $k = 1$, $k' = 2k + 1 = 5$ and $\{a\}$ is an independent set of size $k$ in the graph $G$ from Figure 9. The score for every node is given by it's label.

**Claim 7.** *$G$ has an independent set of size at least $k$ if and only if $D$ has a $k'$-\Circled{O}-neighbor $D'$ with a higher score than $D$ where $k' = 2k + 1$ if Rev $\in \mathbb{O}$ and $k' = 2(2k + 1)$ otherwise.*

We now show how to alter the above construction to obtain the result for maximum degree 3. The only nodes in $S_f$ whose degree may exceed 3 are the nodes in $\{v^1 : v \in V(G)\}$. The main idea to reduce the degree of these nodes is to further split their sets of neighbors using binary trees. For our new construction we define a DAG $D'$ and a local score function $f'$ as follows. The DAG $D'$ is obtained from $D$ by applying the following actions:

1. For every $v \in V(G)$, we delete all nodes of $T_{v^1}$ and all arcs incident with these nodes.

2. For every $v \in V(G)$, we add the nodes $v^{1a}$ and $v^{1b}$ and the arcs $(v^{1a}, v^{1b})$ and $(v^{1b}, v^2)$.

3. For every $v \in V(G)$ we add a binary tree $T_{v^{1a}}$ with exactly $|N_G(v)| + 1$ leaves. The root of $T_{v^{1a}}$ is $v^{1a}$ and all arcs of $T_{v^{1a}}$ are directed away from $v^{1a}$. Furthermore, we define $l_{v^{1a}}$ to be a bijective mapping from $N_G(v) \cup \{l_{T_2}(v)\}$ to the leaves of $T_{v^{1a}}$.

4. For every $v \in V(G)$ we add the arcs $(v^{1b}, l_{T_1}(v))$ and $(l_{v^{1a}}(l_{T_2}(v)), l_{T_2}(v))$.

This completes the construction of $D'$. Next we define the local score function $f'$ as follows:

1. For every $n \in V(D') \setminus \{v^{1a} : v \in V(G)\} \cup \{r_1\}$ we set $f'(n, P_{D'}(n)) = \beta$.

2. For every $v \in V(G)$ we set $f'(v^{1a}, \{v^{1b}\}) = \varepsilon$, $f'(v^{1b}, \{v^2, l_{T_1}(v)\}) = \beta$, $f'(v^2, P_{D'}(v^2) \setminus \{v^{1b}\}) = \beta$, and $f'(l_{T_1}(v), P_{D'}(l_{T_1}(v)) \setminus \{v^{1b}\}) = \beta$.

3. We set $f'(r_1, \mathrm{P}_{D'}(r_1)) = \alpha$ and $f'(r_2, \mathrm{P}_{D'}(r_2) \cup \{r_1\}) = \beta$.

4. For all the remaining combinations of $n \in V(D')$ and $P \subseteq V(D')$ we set $f'(n, P) = 0$.

It is easy to see that $S_{f'}$ has maximum degree 3. Furthermore, using the same arguments as in the proof of Claim 7 one can show that the graph $G$ has an independent set of size $k$ if and only if the DAG $D'$ has a $k'$-⊙-neighbor $D''$ with a higher score than $D'$ (with respect to $f'$) where $k' = 3k + 1$ if REV $\in \mathbb{O}$ and $k' = 2(3k + 1)$ otherwise. □

Theorem 6 provides a surprising contrast to a similar study of $k$-local search for MAX-SAT where the problem is fixed-parameter tractable for instances of bounded degree (Szeider, 2011). A possible explanation for the surprising hardness of $k$-⊙-Local Search Bayesian Structure Learning could be that, in contrast to MAX-SAT, a global property of the entire instance (acyclicity) must be checked.

## 8. The Directed Super-structure

In the previous sections we considered the problem of Exact BNSL and $k$-Local Search BNSL under certain restrictions of the undirected super-structure. However, every strictly admissible solution to Exact BNSL is actually contained in the more restrictive directed super-structure. It is a natural question whether the additional information entailed in the directed super-structure can be used to find new structural restrictions under which Exact BNSL becomes tractable. It is well-known that Exact BNSL becomes significantly easier if an ordering of the variables is given in advance. For instance, given an ordering of the variables of a BN, Exact BNSL becomes solvable in polynomial time if the input is given in the arity-$c$ representation (Teyssier & Koller, 2005). Fixing an ordering of the variables in a BN corresponds to restricting the search space to acyclic directed super-structures. Our first observation of this section is that Exact BNSL is solvable in polynomial time if the directed super-structure is a dag and the input to the problem is given in the more general non-zero representation. It is important to note that there is no corresponding restriction of the undirected super-structure, because restricting the directed super-structure to be acyclic does not impose any restrictions on the undirected super-structure. Considering this promising result it becomes natural to ask whether it is possible to gradually generalize the class of acyclic directed super-structures. A natural approach would be to consider only directed super-structures that can be made acyclic by deleting a small number $k$ of nodes. Such an approach looks promising as it is known that for every fixed $k$ the directed super-structures that can be made acyclic by deleting at most $k$ nodes can be recognized efficiently (Chen, Liu, Lu, O'Sullivan, & Razgon, 2008). However, we show that this approach is unlikely to work for Exact BNSL. Furthermore, in correspondence to the results in the previous sections, NP-hardness even holds if we additionally bound the maximum in-degree and out-degree of $S_{f}^{\rightarrow}$.

**Theorem 7.** Let $N$ be a set of nodes and $f$ a local score function on $N$ such that $S_{f}^{\rightarrow}$ is acyclic. Then we can find in time $O(|N|\delta_f)$ a dag $D$ with maximal score $f(D)$.

*Proof.* Because $S_{f}^{\rightarrow}$ is acyclic, it follows that every strictly admissible directed graph $D$ is also acyclic. Hence in order to compute a dag $D$ with the highest score, it is sufficient to compute for every $n \in N$ a parent set with the highest score. This can clearly be done in time $O(|N|\delta_f)$ and so the result follows. $\qquad\square$

**Corollary 2.** Exact Bayesian Network Structure Learning *is solvable in quadratic time for acyclic directed super-structures.*

*Proof.* This follows immediately from Theorem 7 because both $N$ and $\delta_f$ are bounded by the total input size of the problem. Recall from Section 3 that the local score function $f$ is given as the list of all tuples for which $f$ is non-zero and hence $\delta_f$ is bounded by the total input size of the instance. $\qquad\square$

**Theorem 8.** Exact Bayesian Network Structure Learning *is NP-hard for instances where $S_{\vec{f}}^{\rightarrow}$ can be made acyclic by deleting one node. Hardness even holds if we additionally bound the maximum in-degree and the maximum out-degree of $S_{\vec{f}}^{\rightarrow}$ by 3.*

*Proof.* We devise a polynomial reduction from the restricted version of 3-SAT where every literal is contained in at most two clauses. This version of 3-SAT is still NP-complete (Garey & Johnson, 1979). Let $\phi$ be a 3-CNF formula with variables $x_1, \ldots, x_n$ and clauses $C_1, \ldots, C_m$ where $C_j = l_{j,1} \lor l_{j,2} \lor l_{j,3}$, for every $1 \le j \le m$. We construct a set $N$ of nodes, a local score function $f$ and a real number $s > 0$.

$N$ contains the nodes $d_0, \ldots, d_n$ and $t_0, \ldots, t_{n+m}$ and additionally:

- For every $1 \le i \le n$ the nodes $x_i, \overline{x}_i, a_i, b_i$.

- For every $1 \le j \le m$ the nodes $l_{j,1}, l_{j,2}, l_{j,3}, C_j$.

Let $\alpha = n + 1$ and $\varepsilon = 1$. We define $f$ as follows:

- We set $f(d_0, \{t_0\}) = f(t_0, t_1) = \alpha$.

- For every $1 \le i \le n$ we set:

  - $f(d_i, \{d_{i-1}\}) = \alpha$.
  - $f(a_i, \{d_i\}) = \alpha$ and $f(x_i, \{a_i\}) = f(\overline{x}_i, \{a_i\}) = \varepsilon$.
  - $f(b_i, \{x_i\}) = f(b_i, \{\overline{x}_i\}) = \alpha$.
  - $f(t_i, \{t_{i+1}, b_i\}) = \alpha$.

- For every $1 \le j \le m$ we set:

  - $f(C_j, \{l_{j,1}\}) = f(C_j, \{l_{j,2}\}) = f(C_j, \{l_{j,3}\}) = \alpha$.
  - $f(t_{n+j}, \{t_{n+j+1}, C_j\}) = \alpha$ if $j < m$ and $f(t_{n+j}, \{C_j\}) = \alpha$ if $j = m$.

- For every $1 \le i \le n$, $1 \le j \le m$ and $1 \le l \le 3$ we set $f(l_{j,l}, \{x_i\}) = \alpha$ if $l_{j,l} = x_i$ and $f(l_{j,l}, \{\overline{x}_i\}) = \alpha$ if $l_{j,l} = \overline{x}_i$.

For all other combinations of $v \in N$ and $P \subseteq N$ we set $f(v, P) = 0$. Furthermore, we set $s = (4n + 5m + 2)\alpha + n\varepsilon$. An example of the directed super-structure constructed from a 3-CNF formula is shown in Figure 11. We establish the theorem by showing the following claims.

**Claim 8.** $S_{\vec{f}}$ can be made acyclic by deleting at most one node.

**Claim 9.** $S_{\vec{f}}$ has maximum in-degree and maximum out-degree 3.

**Claim 10.** $\phi$ is satisfiable if and only if there is a dag $D$ with score $f(D) \geq s$.

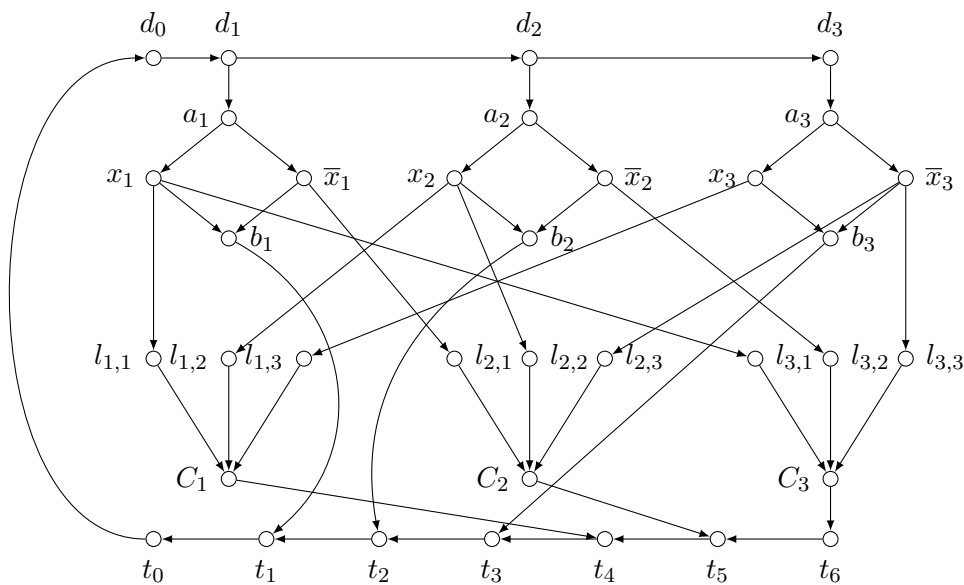

Figure 11: The directed super-structure for the formula $\phi = (x_1 \vee x_2 \vee x_3) \wedge (\overline{x}_1 \vee x_2 \vee \overline{x}_3) \wedge (x_1 \vee \overline{x}_2 \vee \overline{x}_3)$ in the proof of Theorem 8.

It is easy to see that $S_{\vec{f}} - d_0$ is acyclic and hence Claim 8 holds. Since every literal occurs in at most two clauses it is also easy to verify Claim 9. The proof of Claim 10 is straightforward and can be found in the appendix. □

## 9. Conclusion

We have studied the computational complexity of Bayesian Structure Learning (BNSL) under various restrictions on the (directed) super-structure, considering both Exact BNSL and $k$-Local Search BNSL. We have obtained positive and negative results for the theoretical worst-case complexities of the problems. Our main positive result states that Exact BNSL is linear-time tractable if the super-structure has bounded treewidth and bounded maximum degree. We have contrasted out positive results with negative results, using techniques and concepts from Parameterized Complexity. This theoretical framework is particularly well-suited for such an investigation as it allows a fined-grained investigation that takes structural aspects of problem instances into account. Our results point out which combinations of structural restrictions make the problems tractable and which restrictions cannot be dropped without loosing tractability. Considering various combinations of restrictions in a systematic way allows us draw a broader picture of the complexity landscape (see the table in Section 1). We hope that our results provide a better understanding of the principles of BNSL and contribute to its foundations. We hope that this understanding will also be useful for the development of heuristic methods for practical BNSL systems.

## Acknowledgments

A shorter and preliminary version of this paper appeared in UAI 2010. Research supported by the European Research Council, grant reference 239962.

## Appendix A

*Proof of Claim 1 (Lemma 2).* Let us first assume that $R = (a, p, s)$ is a valid record of $t$ and let $P = a(v_0)$. Since $R$ is valid it follows that it represents some partial solution $D \in \mathcal{D}(t)$ such that $D$ is maximal with respect to $R$. Now, $D' = D[\chi^*(t')]$ is a partial solution for $t'$ and it follows from Proposition 4 that $D' = D - v_0$ and furthermore $f_{t'}(D') = s$. Hence, $D'$ can be represented by some record $R' = (a', p', s')$ such that $R$, $R'$ and $P$ satisfy the conditions of this claim. Furthermore, since $R$ is valid and $f_t(D) = f_{t'}(D')$ the maximality of $f_{t'}(D')$ with respect to $R'$ follows from the maximality of $f_t(D)$ with respect to $R$ and hence $R'$ is a valid record of $t'$.

To see the converse let $P \in \mathcal{P}_f(v_0)$ and a valid record $R' = (a', p', s') \in \mathcal{R}(t')$ be given. Let $R = (a, p, s)$ be the triple as defined via the conditions (1)-(4). Since $R'$ is a valid record it represents some partial solution $D'$ such that $D'$ is maximal with respect to $R'$. It is easy to see that the digraph $D$ with node set $V(D') \cup \{v_0\}$ and arc set $E(D') \cup \{(u, v_0) : u \in P\} \cup \{(v_0, u) : u \in \chi(t')$ such that $v_0 \in a'(u)\}$ is acyclic if and only if $p$ satisfies condition (5), i.e., $p$ is irreflexive. It follows that $R$ represents $D$ if and only if $p$ satisfies condition (5) and furthermore the maximality of $D$ with respect to $R$ follows from the maximality of $D'$ with respect to $R'$. Hence, $R$ is a valid record for $t$ if and only if it satisfies the conditions (1)–(5). □

*Proof of Claim 2 (Lemma 3).* Let us first assume that $R = (a, p, s)$ is a valid record of $t$. Since $R$ is a valid record it represents some solution $D \in \mathcal{D}(t)$ such that $D$ is maximal with respect to $R$. Now, let $R' = (a', p', s')$ be such that $a'(v_0) = P_D(v_0)$, $a'(v) = P_D(v)$ for every $v \in \chi(t)$, $p'$ is the union of $p$ and all tuples $(v_0, v)$ and $(v, v_0)$ where $v \in \chi(t)$ such that there is a directed path from $v_0$ to $v$ respectively from $v$ to $v_0$ in $D$ and $s' = s - f(v_0, P_D(v_0))$. Note that because of Proposition 5 we can assume that $P_D(v_0) \in \mathcal{P}_f(v_0)$ and hence $D$ is also represented by $R'$. It follows from the maximality of $D$ with respect to $R$ that $R'$ is a maximal element in $\mathcal{R}(t')[t]$.

To see the converse let $R = (a, p, s)$ be a maximal element in $\mathcal{R}(t')[t]$. Since $R \in \mathcal{R}(t')[t]$ it follows that there is a record $R' = (a', p', s') \in \mathcal{R}(t')$ such that $R = R'[t]$. Hence, there is a partial solution $D$ represented by $R'$ such that $f_{t'}(D) = s'$ is maximal with respect to all partial solutions represented by $R'$. Clearly, $D$ is also represented by $R$ and the maximality of $D$ with respect to $R$ follows from the fact that $R$ is a maximal element in $\mathcal{R}(t')[t]$. □

*Proof of Claim 3 (Lemma 4).* Let us first assume that $R = (a, p, s)$ is a valid record for $t$. Since $R$ is valid it follows that it represents a partial solution $D \in \mathcal{D}(t)$ such that $D$ is maximal with respect to $R$. Let $D_1 = D[\chi^*(t_1)]$ and $D_2 = D[\chi^*(t_2)]$, i.e., $D_1$ and $D_2$ are the two subdigraphs of $D$ induced by the nodes contained in the nodes of the subtrees rooted at $t_1$ and $t_2$, respectively. It follows from Proposition 3 that $D = D_1 \cup D_2$ and $V(D_1) \cap V(D_2) = \chi(t)$. Hence, $D_1$ and $D_2$ are partial solutions for $t_1$ and $t_2$, respectively. For $i \in \{1, 2\}$, let $R_i = (a_i, p_i, s_i)$ be such that $a = a_i$, $(v, w) \in p_i$ if and only if there is a directed path from $v$ to $w$ in $D_i$ and $s_i = f_{t_i}(D_i)$. It follows directly from the definition that $R_1$ and $R_2$ represent $D_1$ and $D_2$, respectively, and since $f_t(D) = f_{t_1}(D_1) + f_{t_2}(D_2)$ the maximality of $D$ with respect to $R$ implies the maximality of $D_1$ and $D_2$ with respect to $R_1$ and $R_2$, respectively. Hence, $R_1$ and $R_2$ are valid records for $t_1$ and $t_2$, respectively, and it is easy to see that $R$, $R_1$ and $R_2$ satisfy the conditions of the claim.

To see the converse let us assume that we are given $R_1 = (a_1, p_1, s_1) \in \mathcal{R}(t_1)$ and $R_2 = (a_2, p_2, s_2) \in \mathcal{R}(t_2)$ and that the triple $R = (a, p, s)$ as defined by the conditions (1)–(3) satisfies condition (4). Since $R_1$ and $R_2$ are valid it follows that both represent some partial solutions $D_1$ and $D_2$ such that $D_1$ and $D_2$ are maximal with respect to $R_1$ and $R_2$, respectively. Furthermore, using Proposition 3 it follows that $V(D_1) \cap V(D_2) = \chi(t)$ and hence it follows from condition (4) that $D = D_1 \cup D_2$ is a partial solution represented by $R$. Now, $f_t(D) = f_{t_1}(D_1) + f_{t_2}(D_2) = s_1 + s_2 = s$ and the maximality of $D$ with respect to $R$ follows from the maximality of $D_1$ and $D_2$ with respect to $R_1$ and $R_2$, respectively. It follows that $R$ is a valid record for $t$. $\qquad\square$

*Proof of Claim 6 (Theorem 3).* **(1)$\Rightarrow$(2)** Suppose $G$ has a $k$-clique $K$. Then $f(D(E(K))) = nk\alpha - |V(K)|\alpha + |E(K)|\varepsilon = nk\alpha + k \geq s$ and it remains to show that $D(E(K))$ is acyclic. To see this note that every cycle in $D(E(K))$ has to use at least one node from $V$, because $D(E(K))$ does not contain an arc between two nodes in $A$. But since $K$ is a clique, every node $v \in V$ is either a sink, i.e., $v$ has only incoming arcs, or a source, i.e., $v$ has only outgoing arcs and hence no cycle can use a node from $V$.

**(2)$\Rightarrow$(3)** Suppose $D$ is a dag with $f(D) \geq s$. Let $A'$ be the set of nodes in $A$ with $f(a, P_D(a)) = \varepsilon$ for every $a \in A'$. It follows from the definition of $f$ that for every $a_{ij} \in A'$ there is a unique edge $e$ between a node in $V_i$ and a node in $V_j$ in $G$ with $P_D(a_{ij}) = e$. Let $E' \subseteq E(G)$ be the set of all edges in $G$ that correspond to a node in $A'$. It follows that $E'$ is representable and we claim that every node $v \in N$ has at least the local score in $D(E')$ as it has in $D$. By construction of $D(E')$ the claim is trivially satisfied for every node $a \in A$. Similarly, for every $v \in V \setminus V(E')$ it holds that $f(v, P_{D(E')}(v)) = \alpha$ and hence $f(v, P_{D(E')}(v)) \geq f(v, P_D(v))$. Furthermore, for every $v \in V(E')$ it holds that $f(v, P_D(v)) = 0$ and hence again $f(v, P_{D(E')}(v)) \geq f(v, P_D(v))$. It follows that $f(D(E')) \geq f(D) \geq s$.

**(3)$\Rightarrow$(1)** Suppose that $E' \subseteq E(G)$ is a representable edge set of $G$ with $f(D(E')) \geq s$. We will show that $|V(E')| = k$ and $|E'| = \binom{k}{2}$ which implies that $E'$ is the edge set of a $k$-clique in $G$.

Because $E'$ is an edge set, it holds that $|E'| \leq \binom{|V(E')|}{2}$ and hence:

$$
\begin{aligned}
f(D(E')) - nk\alpha &= -|V(E')|\alpha + |E'|\varepsilon \\
&\leq -|V(E')|\alpha + \binom{|V(E')|}{2}\varepsilon \\
&= -|V(E')|(k^2 - k - 1) + \binom{|V(E')|}{2}2k \\
&= -|V(E')|(k^2 - k - 1) + (|V(E')|^2 - |V(E')|)k \\
&= -|V(E')|k^2 + |V(E')|^2k + |V(E')|k - |V(E')|k + |V(E')| \\
&= -|V(E')|k^2 + |V(E')|^2k + |V(E')|
\end{aligned}
$$

Because $f(D(E')) \geq s$, it follows that $-|V(E')|k^2 + |V(E')|^2k + |V(E')| \geq 1$ and hence:

$$
\begin{aligned}
-|V(E')|k^2 + |V(E')|^2k + |V(E')| &\geq 1 \\
-k^2 + |V(E')|k + 1 &\geq \frac{1}{|V(E')|} \\
|V(E')| &\geq \frac{\frac{1}{|V(E')|} + k^2 - 1}{k} \\
|V(E')| \geq k - \frac{1}{k} + \frac{1}{|V(E')|k}
\end{aligned}
$$

Since $|V(E')|$ is an integer and $k > 2$, we have that $|V(E')| \geq k$.

Furthermore, since $E'$ is representable it can contain at most $\binom{k}{2}$ edges and hence $f(D(E')) - nk\alpha \leq -|V(E')|\alpha + \binom{k}{2}\varepsilon$. Again, it follows from $f(D(E')) \geq s$ that $-|V(E')|\alpha + \binom{k}{2}\varepsilon \geq 1$ and:

$$
\begin{aligned}
-|V(E')|\alpha + \binom{k}{2}\varepsilon &\geq 1 \\
-|V(E')|(k^2 - k - 1) + k^3 - k^2 &\geq 1 \\
-|V(E')|(k^2 - k - 1) &\geq -k^3 + k^2 + 1 \\
-|V(E')| &\geq \frac{-k^3 + k^2 + 1}{k^2 - k - 1} \\
|V(E')| &\leq k + \frac{k + 1}{k^2 - k - 1}
\end{aligned}
$$

Since $|V(E')|$ is an integer and $k > 2$, it follows that $|V(E')| \leq k$ and hence $|V(E')| = k$. Using this in $f(D(E')) - nk\alpha \geq 1$ we get:

$$
\begin{aligned}
-k\alpha + |E'|\varepsilon &\geq 1 \\
-(k^3 - k^2 - k) + |E'|2k &\geq 1 \\
|E'| &\geq \frac{k^3 - k^2 - k + 1}{2k} \\
|E'| &\geq \frac{k^2 - k - 1 + \frac{1}{k}}{2} \\
|E'| &\geq \binom{k}{2} - \frac{1 - \frac{1}{k}}{2}
\end{aligned}
$$

Because $|E'|$ is an integer and $k > 2$, it follows that $|E'| \geq \binom{k}{2}$. Putting everything together we have that $f(D(E')) \geq s$ implies $|V(E')| = k$ and $|E'| = \binom{k}{2}$. Hence $E'$ is the edge set of a $k$-clique in $G$. $\qquad \square$

*Proof of Claim 7 (Theorem 6).* We first show the claim for the case that REV $\in \mathbb{O}$ and $k' = 2k + 1$.

Let us first assume that $G$ has an independent set $S \subseteq V(G)$ of size at least $k$. We obtain $D'$ from $D$ by reversing the $k'$ arcs in $\{ (v^1, v^2), (v^1, l_{T_1}(v)) : v \in S \} \cup \{ (r_2, r_1) \}$. This decreases the score for $r_1$ by $\alpha$ and increases the score for the nodes in $\{ v^1 : v \in S \}$ by $\varepsilon$ while the score of all the other nodes of $D$ remains unchanged. Hence, $f(D') = f(D) - \alpha + |S|\varepsilon \geq \alpha + k\varepsilon = f(D) + 1 > f(D)$ and it remains to show that $D'$ is acyclic. To see this assume that $D'$ contains a cycle $C$. Because $D$ is acyclic $C$ must contain at least one of the newly created arcs in $D'$, i.e., $C$ must contain either an arc $(v^2, v^1)$, an arc $(l_{T_1(v)}, v^1)$ for some $v \in S$ or the arc $(r_1, r_2)$. Because $r_2$ is a sink in $D'$, i.e., $r_2$ has no outgoing arcs, the cycle $C$ cannot contain the arc $(r_1, r_2)$. Similarly, because $D'$ does not contain a directed path from $v^1$ to $l_{T_1}(v)$ the cycle $C$ cannot contain an arc $(l_{T_1(v)}, v^1)$ for any $v \in S$. Hence the cycle $C$ must contain an arc $(v^2, v^1)$ for some $v \in S$. So suppose that $C$ contains the arc $(v^2, v^1)$ for some $v \in S$. Because $D'$ contains no directed paths from a node of $T_2$ to a node of $T_1$ it follows that $C$ cannot leave the node $v^1$ using the arc $(v^1, l_{T_2}(v))$. Consequently, the cycle $C$ must leave the node $v^1$ towards $w^2$ for some neighbor $w$ of $v$ in $G$. Because $S$ is an independent set in $G$ it follows that $w \notin S$ and hence the node $w^2$ is a sink in $D'$ contradicting the existence of the cycle $C$.

To see the reverse direction assume that $D'$ is a dag obtained from $D$ by reversing at most $k'$ arcs. Note that the nodes in $\{ v^1 : v \in V(G) \}$ are the only nodes of $D$ whose scores are not yet maximum. Hence, in order for $D'$ to have a higher score than $D$ the score for at least one of these nodes has to be increased. Now, the score for such a node $v^1$ for some $v \in V(G)$ can only be increased by reversing the arcs $(v^1, v^2)$ and $(v^1, l_{T_1}(v))$. It is easy to see that reversing the arc $(v^1, l_{T_1}(v))$ introduces a cycle $C$ in $D$ that uses only nodes in $V(T_1) \cup V(T_2) \cup \{v^1\}$. However, because every such cycle $C$ contains the arc $(r_2, r_1)$ we can destroy all of these cycles by additionally reversing the arc $(r_2, r_1)$. Because reversing $(r_2, r_1)$ does only decrease the score of $r_1$ by $\alpha$ this is also the cheapest way to destroy these cycles. Now, $\alpha = (k-1)\varepsilon$ and it follows that in order to increase the score of $D$ the scores of at least $k$ nodes in $\{ v^1 : v \in V(G) \}$ have to be increased to $\varepsilon$. Let $S$ be the set of nodes in $V(G)$ such that the score of the nodes in $\{ v^1 : v \in S \}$ has been increased in this manner. As mentioned above $|S| \geq k$ and it remains to show that $S$ is an independent set in $G$. Suppose $S$ is not an independent set and let $u, v \in S$ be such that $\{u, v\} \in E(G)$. Then the arcs $(v^2, v^1)$ and $(u^2, u^1)$ together with the directed path from $v^1$ to $u^2$ (using the arcs in $T_{v^1}$ and $T_{u^2}$) and the directed path from $u^1$ to $v^2$ (using the arcs in $T_{u^1}$ and $T_{v^2}$) form a cycle in $D'$ contradicting the acyclicity of $D'$. It follows that $S$ is an independent set in $G$ of size at least $k$.

Hence, we have shown the theorem for the case REV $\in \mathbb{O}$. It remains to show the theorem for the only remaining non-trivial set with REV $\notin \mathbb{O}$, i.e., the set $\mathbb{O} = \{\text{ADD}, \text{DEL}\}$. Now $k' = 2(2k + 1)$ and the idea is to replace every reversal of an arc $(u, v)$ by a deletion (of $(u, v)$) and an addition (of $(v, u)$). $\qquad \square$

*Proof of Claim 10 (Theorem 8).* We will prove Claim 10 with the help of the following claim.

**Claim 11.** $f(D) \geq s$ and $D$ is acyclic if and only if $D$ satisfies the following conditions:

1 *$D$ contains at least the following arcs:*

- *The arc $(t_0, d_0)$.*
- *For every $1 \leq i \leq n$, the arcs $(d_{i-1}, d_i)$, $(d_i, a_i)$, $(b_i, t_i)$ and $(t_i, t_{i-1})$.*
- *For every $1 \leq i \leq n$, $1 \leq j \leq m$ and $1 \leq l \leq 3$ the arc $(x_i, l_{j,l})$ if $l_{j,l} = x_i$ and similarly the arc $(\overline{x}_i, l_{j,l})$ if $l_{j,l} = \overline{x}_i$.*
- *For every $1 \leq j \leq m$, the arcs $(C_j, t_{n+j})$ and $(t_{n+j}, t_{n+j-1})$ and exactly one of the arcs $(l_{j,1}, C_j)$, $(l_{j,2}, C_j)$ and $(l_{j,3}, C_j)$.*

2 *For every $1 \leq i \leq n$ the digraph $D$ contains the arcs $(a_i, x_i)$ and $(\overline{x}_i, b_i)$ but not the arcs $(a_i, \overline{x}_i)$ and $(x_i, b_i)$ or $D$ contains the arcs $(a_i, \overline{x}_i)$ and $(x_i, b_i)$ but not the arcs $(a_i, x_i)$ and $(\overline{x}_i, b_i)$.*

3 *For $1 \leq i \leq n$, $1 \leq j \leq m$ and $1 \leq l \leq 3$, the following holds:*

- *If $l_{j,l} = x_i$ and $D$ contains the arc $(l_{j,l}, C_j)$ then $D$ does not contain the arc $(a_i, x_i)$.*
- *If $l_{j,l} = \overline{x}_i$ and $D$ contains the arc $(l_{j,l}, C_j)$ then $D$ does not contain the arc $(a_i, \overline{x}_i)$.*

We will first show how the previous claim can be used to prove Claim 10. Suppose $\phi$ is satisfiable and let $\beta$ be a satisfying assignment for $\phi$. Let $D$ be the digraph that satisfies condition *1* and additionally:

- For every $1 \leq i \leq n$ if $\beta(x_i) = \text{true}$ then $D$ contains the arcs $(a_i, \overline{x}_i)$ and $(x_i, b_i)$, otherwise $D$ contains the arcs $(a_i, x_i)$ and $(\overline{x}_i, b_i)$.

- For every $1 \leq j \leq m$ let $l_{j,l}$ any literal in the clause $C_j$ that is satisfied by $\beta$; since $\beta$ is a satisfying assignment every clause $C_j$ contains such a literal. Then $D$ contains the arc $(l_{j,l}, C_j)$.

It follows that $D$ satisfies the conditions *2* and *3* and hence (using Claim 11) $f(D) \geq s$ and $D$ is acyclic.

To see the reverse let $D$ be a dag with $f(D) \geq s$. It follows from Claim 11 that $D$ satisfies the conditions *1–3*. We claim that the assignment $\beta$ with $\beta(x_i) = \text{true}$ if and only if $D$ does not contain the arc $(a_i, x_i)$ is a satisfying assignment for $\phi$. It follows from condition *1* that for every $1 \leq j \leq m$ the digraph $D$ contains an arc $(l_{j,l}, C_j)$ for some $1 \leq l \leq 3$. W.l.o.g., we can assume that $l_{j,l} = x_i$ for some $1 \leq i \leq n$ (the case that $l_{j,l} = \overline{x}_i$ is analog). Again using condition *1* it follows that $D$ contains the arc $(x_i, l_{j,l})$. Because of condition *3* the digraph $D$ does not contain the arc $(a_i, x_i)$ and hence $\beta(l_{j,l}) = \text{true}$.

Hence it only remains to show Claim 11. Let us first show that every dag $D$ with $f(D) \geq s$ satisfies conditions *1–3*. To see this observe that every node in $V = \{x_1, \overline{x}_1, \ldots, x_n, \overline{x}_n\}$ has either score $0$ or $\varepsilon$. Similarly, every node in $V' = N \setminus V$ has either score $0$ or $\alpha$. It follows that in every directed graph there are at most $4n + 5m + 2$ nodes with score $\alpha$ and at most $2n$ nodes with score $\varepsilon$. Hence, the maximum score for every directed graph is $(4n + 5m + 2)\alpha + 2n\varepsilon$. Because $\alpha > n\varepsilon$ and $f(D) \geq s$ it follows that in $D$ every node from $V'$ must have score $\alpha$ and similarly at least $n$ of the $2n$ nodes in $V$ must have score $\varepsilon$. Hence $D$ satisfies condition 1.

To show condition 2 observe that because for every $1 \leq i \leq n$ the node $b_i$ must have score $\alpha$ in $D$ it holds that exactly one of the arcs $(x_i, b_i)$ and $(\overline{x}_i, b_i)$ is in $D$. Now, if $D$ contains the arc $(x_i, b_i)$ for some $1 \leq i \leq n$ then $D$ cannot contain the arc $(a_i, x_i)$ because otherwise $D$ would contain the cycle $(d, a_i, x_i, b_i, t, d)$. Similarly, if $D$ contains the arc $(\overline{x}_i, b_i)$ for some $1 \leq i \leq n$ then $D$ cannot contain the arc $(a_i, \overline{x}_i)$. It follows that for every $1 \leq i \leq n$ at least one of the arcs $(a_i, x_i)$ and $(a_i, \overline{x}_i)$ is missing in $D$. Since there are at most $n$ nodes in $V$ with score $0$ it follows that for every $1 \leq i \leq n$ exactly one of the arcs $(a_i, x_i)$ and $(a_i, \overline{x}_i)$ must be in $D$. It follows that $D$ satisfies condition 2.

To see condition 3 suppose for some $1 \leq i \leq n$, $1 \leq j \leq m$ and $1 \leq l \leq 3$ with $l_{j,l} = x_i$ the digraph $D$ contains both arcs $(l_{j,l}, C_j)$ and $(a_i, x_i)$. It follows that $D$ would contain the cycle $(d, a_i, x_i, l_{j,l}, C_j, t, d)$, a contradiction. The case that $l_{j,l} = \overline{x}_i$ is analog and hence $D$ satisfies condition 3.

To see the reverse implication of the claim suppose that $D$ is a digraph that satisfies the conditions 1–3. It is easy to see that $f(D) = s$ and it hence only remains to show that $D$ is acyclic. Because $S_{\vec{f}} - (t_0, d_0)$ is acyclic it follows that every cycle in $D$ has to use the arc $(t_0, d_0)$. Hence $D$ contains a cycle if and only if there is a directed path $P$ from $d_0$ to $t_0$ in $D$. It follows from condition 2 that there is no directed path from $d_0$ to some $b_i$ in $D$, for $1 \leq i \leq 3$, and hence $P$ cannot contain a node $b_i$. Since the only other nodes in $D$ with arcs to $\{t_0, \ldots, t_{n+m}\}$ are the nodes $C_1, \ldots, C_m$ it follows that $P$ has to use a node $C_j$ for some $1 \leq j \leq m$. Because of condition 1 the node $C_j$ has exactly one incoming neighbor (one of $l_{j,1}, l_{j,2}, l_{j,3}$) say $l_{j,l}$. Again using condition 1 the node $l_{j,l}$ has exactly one incoming neighbor $x_i$ or $\overline{x}_i$ if $l_{j,l} = x_i$ or $l_{j,l} = \overline{x}_i$, respectively. W.l.o.g. let us assume that $l_{j,l} = x_i$. It follows from condition 3 that $x_i$ has no incoming neighbor and hence $D$ contains no directed path $P$ from $d_0$ to $t_0$. $\qquad\square$